\pdfoutput=1

\documentclass[11pt]{article}

\usepackage{acl}

\usepackage{times}
\usepackage{latexsym}

\usepackage[T1]{fontenc}

\usepackage[utf8]{inputenc}

\usepackage{microtype}

\usepackage{inconsolata}

\usepackage{multirow}
\usepackage{makecell}

\usepackage{times}
\usepackage{latexsym}
\usepackage{graphicx}
\usepackage{subfigure}
\usepackage{float}
\graphicspath{{images}}
\usepackage{stfloats}
\usepackage{booktabs}
\usepackage{indentfirst}
\usepackage{amsmath}
\usepackage{amssymb}

\usepackage{enumitem}

\usepackage{colortbl}
\usepackage{arydshln}

\usepackage[linesnumbered,ruled,vlined]{algorithm2e}

\SetCommentSty{mycommfont}

\SetKwInput{KwInput}{Input}                
\SetKwInput{KwOutput}{Output}              

\usepackage{CJKutf8}

\usepackage{listings}
\usepackage{xcolor}

\definecolor{codegreen}{rgb}{0,0.6,0}
\definecolor{codegray}{rgb}{0.5,0.5,0.5}
\definecolor{codepurple}{rgb}{0.58,0,0.82}
\definecolor{backcolour}{rgb}{0.95,0.95,0.92}

\lstdefinestyle{mystyle}{
    backgroundcolor=\color{backcolour},   
    commentstyle=\color{codegreen},
    keywordstyle=\color{magenta},
    numberstyle=\tiny\color{codegray},
    stringstyle=\color{codepurple},
    basicstyle=\ttfamily\small,
    breakatwhitespace=false,         
    breaklines=true,                 
    captionpos=b,                    
    keepspaces=true,                 
    numbers=left,                    
    numbersep=5pt,                  
    showspaces=false,                
    showstringspaces=false,
    showtabs=false,                  
    tabsize=2
}

\title{MRAG: Benchmarking Retrieval-Augmented Generation for Bio-medicine}

\author{
    Liz Li\textsuperscript{1,}, Wei Zhu\textsuperscript{2,}\thanks{\ \ Corresponding author. For any inquiries, please contact: michaelwzhu91@gmail.com. }\\
    \small \textsuperscript{1}\ DataSelect AI, Xuhui, Shanghai, China \\
    \small \textsuperscript{2}University of Hong Kong, Hong Kong, HK, China
}

\begin{document}
\maketitle
\begin{abstract}

While Retrieval-Augmented Generation (RAG) has been swiftly adopted in scientific and clinical QA systems, a comprehensive evaluation benchmark in the medical domain is lacking. To address this gap, we introduce the Medical Retrieval-Augmented Generation (MRAG) benchmark, covering various tasks in English and Chinese languages, and building a corpus with Wikipedia and Pubmed. Additionally, we develop the MRAG-Toolkit, facilitating systematic exploration of different RAG components. Our experiments reveal that: (a) RAG enhances LLM reliability across MRAG tasks. (b) the performance of RAG systems is influenced by retrieval approaches, model sizes, and prompting strategies. (c) While RAG improves usefulness and reasoning quality, LLM responses may become slightly less readable for long-form questions. We will release the MRAG-Bench's dataset and toolkit with CCBY-4.0 license upon acceptance, to facilitate applications from both academia and industry. 

\end{abstract}

\begin{CJK*}{UTF8}{gbsn}

\section{Introduction}

Large Language Models (LLMs) have transformed how people seek information online, shifting from searching through websites to directly asking chatbots for answers. Recent studies have demonstrated their state-of-the-art capabilities in question answering (QA) across both general and medical domains \cite{achiam2023gpt4,anil2023palm2,singhal2023large,singhal2023towards,nori2023capabilities,huang2023c,li2023cmmlu,Cui2023UltraFeedbackBL,wang2024ts,yue2023-TCMEB,Zhang2023LearnedAA,2023arXiv230318223Z,Xu2023ParameterEfficientFM,Ding2022DeltaTA,Xin2024ParameterEfficientFF,qin2023chatgpt,PromptCBLUE,text2dt_shared_task,Text2dt,zhu_etal_2021_paht,Li2023UnifiedDR,Zhu2023BADGESU,Zhang2023LECOIE,Zhu2023OverviewOT,guo-etal-2021-global,zhu-etal-2021-discovering,Zheng2023CandidateSF,info:doi/10.2196/17653,Zhang2023NAGNERAU,Zhang2023FastNERSU,Wang2023MultitaskEL,Zhu2019TheDS,zhu2021leebert,Zhang2021AutomaticSN,Wang2020MiningIH,li2025ft,leong2025amas,zhang2025time,yin2024machine}. However, LLMs often produce plausible but factually incorrect responses, a phenomenon known as hallucination \cite{ji2023survey,rawte2023survey}. Additionally, the training data for LLMs may not encompass the latest knowledge, such as recent medical literature in PubMed\footnote{\url{https://pubmed.ncbi.nlm.nih.gov/}.} or the latest updates to clinical guidelines. These issues pose significant risks in high-stakes scenarios like healthcare \cite{tian2024opportunities,hersh2024search,zhu2024iapt,zhu-tan-2023-spt,Liu2022FewShotPF,xie2024pedro,Cui2023UltraFeedbackBL,zheng2024nat4at,zhu2023acf,gao2023f,zuo-etal-2022-continually,zhang-etal-2022-pcee,sun-etal-2022-simple,zhu-etal-2021-gaml,Zhu2021MVPBERTMP,li-etal-2019-pingan,zhu2019panlp,zhu2019dr,zhou2019analysis,zhang2025time,wang2025ts,liu2025parameter,yi2024drum,tian2024fanlora}.

Retrieval-Augmented Generation (RAG) leverages up-to-date and reliable document collections to enhance the capabilities of Large Language Models (LLMs), potentially resolving various challenges in the field \cite{lewis2020retrieval,gao2023retrieval,zhao2024retrieval}. By grounding the reasoning of LLMs in the retrieved documents, RAG also enhances their transparency. Consequently, RAG has rapidly been adopted in numerous scientific and clinical question-answering systems \cite{lala2023paperqa,jin2024agentmd,zakka2024almanac}. A complete RAG system comprises several flexible modules, including document collections (corpora), retrieval algorithms (retrievers), and backbone LLMs. Given the diverse tasks within the medical domain, RAG's roles can vary significantly. Therefore, a comprehensive evaluation of RAG in the medical domain is critically important.

We first construct a comprehensive Medical Retrieval-Augmented Generation benchmark (MRAG-Bench) to evaluate the LLM-based RAG systems systematically. MRAG covers 4 task cohorts, two major languages, English and Chinese, and a total of 13 test datasets and 14,816 test samples (see Figure \ref{fig:task_compositions} for visualization of task compositions). We also develop and open-source the MRAG-Toolkit, an off-the-shelf toolkit (see Figure \ref{fig:architecture}) that supports (a) three different retrieval approaches, sparse retrieval, semantic retrieval, and webpage search, (b) different retrieval algorithms or models, (c) various API based or locally deployed LLMs, and (d) different prompt strategies. Extensive experiments are conducted on the MRAG benchmark using the MRAG-Toolkit, which results in the following observations: (a) RAG indeed helps the LLMs to become more reliable on all four types of MRAG tasks. (b) LLMs' performance is directly affected by the referential corpus, the retrieval approaches/models, and the prompting strategies. (c) LLMs' performance is log-linearly related to the model's sizes, and larger LLMs tend to benefit more from RAG. (d) Although benefiting from referential documents regarding reasoning, medical knowledge, and overall usefulness, LLMs' responses become slightly less readable when answering long-form questions.

In summary, our contributions are three-fold:
\begin{itemize}
\item We introduce a comprehensive RAG evaluation benchmark, MRAG-Bench, for large language models in the medical domain. Our benchmark provides a suitable testbed for the academic and industrial RAG systems, especially those focused on the bio-medical domain. 

\item We provide an accompanying toolkit, MRAG-Toolkit, for systematically investigating how different components of the MRAG system affect performance. 

\item We have conducted extensive experiments which reveal how to improve an LLM's performance on MRAG tasks.

\end{itemize}

\section{Related work}

Due to limited length, more related works on bio-medical question answering are presented in Appendix \ref{sec:app_related_works}. 

Retrieval-Augmented Generation (RAG) was proposed by \cite{lewis2020retrieval} to enhance the generation performance on knowledge-intensive tasks by integrating retrieved relevant information. In the LLM era led by OpenAI's ChatGPT and GPT-4, RAG not only mitigates the problem of hallucinations as LLMs are grounded on given contexts but can also provide up-to-date knowledge that the LLMs might not encode \cite{gao2023retrieval,zhao2024retrieval}. Many recent studies have been devoted to improving upon the vanilla RAG workflow by either designing novel retrieval and generation mechanisms \cite{borgeaud2022improving,zhang2023repocoder,ram2023context,jiang2023active}, or incorporating pre-training and fine-tuning for improving LLMs' capabilities in RAG \cite{zhang2024raft,siriwardhana2023improving,xue2024badrag}.

In the bio-medicine domain, current systematic evaluations of LLMs typically focus on the vanilla LLMs without RAG \cite{PromptCBLUE,singhal2023large,singhal2023towards,nori2023capabilities,chen2023large,saab2024capabilities}. There has been a series of works on how RAG can help to improve LLMs' capabilities in tasks like clinical decision-making, literature analysis, and information extraction \cite{frisoni2022bioreader,naik2021literature,xiong2024benchmarking,lala2023paperqa,jin2024agentmd,zakka2024almanac,jeong2024improving,wang2023augmenting}. However, (a) a comprehensive evaluation benchmark that contains a variety of tasks is lacking, and (b) systematic investigations on how to build a RAG system, such as the prompt strategies, in the medical domain is lacking. Our work compliments the existing literature by constructing a comprehensive evaluation benchmark for the LLM-based RAG system.

\section{The MRAG Benchmark}
\label{sec:mrg_bench}

\subsection{Constituting tasks}

\begin{figure}
\centering
\includegraphics[width=0.4\textwidth]{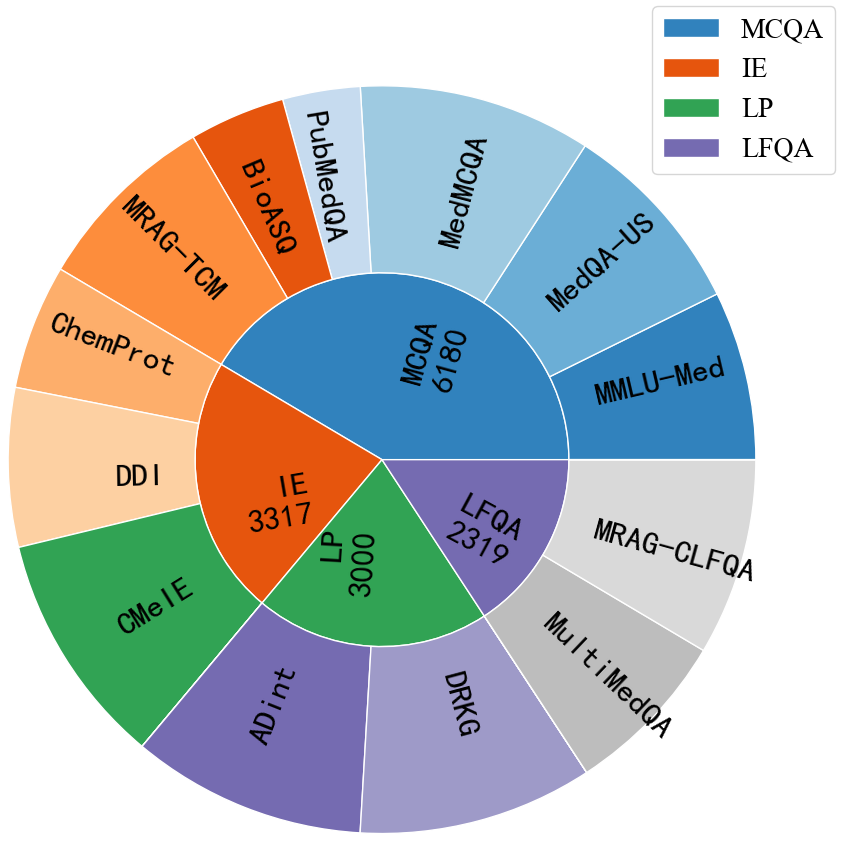}
\caption{Composition of tasks in our MRAG-Bench. }
\label{fig:task_compositions}
\vspace{-8pt}
\end{figure}

To better evaluate LLMs' capabilities in medical RAG, we consider a variety of task types in the medical domain, including multi-choice question answering (MCQA), information extraction (IE), link prediction (LP), and long-form question answering (LFQA). The task compositions are presented as a pie chart in Figure \ref{fig:task_compositions}, where each task's corresponding area is proportional to its test set size.

\textbf{Multi-choice question-answering (MCQA).} To be consistent with existing literature on LLMs' evaluation \cite{hendrycks2020measuring,suzgun2022challenging,wang2023cmb,yue2024tcmbench}, we include five commonly used English MCQA tasks, including three medical examination QA datasets (MMLU-Med \cite{hendrycks2020measuring}, MedQA-US \cite{jin2020disease}, MedMCQA \cite{pal2022medmcqa}) and two biomedical research QA datasets (PubMedQA \cite{jin2019pubmedqa}, BioASQ-Y/N \cite{krithara2023bioasq}).

For Chinese MCQA, since the exam questions of Western medicine are well covered by the MMLU-Med, MedQA-US, and MedMCQA tasks, we construct an MCQA dataset containing 1200 test samples for Traditional Chinese Medicine (TCM) \cite{xue2003studying}, and refer to this dataset as MRAG-TCM. This dataset is a robust benchmark for testing the efficacy and accuracy of language models (with or without RAG) in understanding and generating responses pertinent to TCM.

\textbf{Long-form question answering (LFQA).} To better reflect how online users obtain medical information, we also include long-form question-answering in our MRAG benchmark, in which a question does not have a precise answer like the multi-choice setting. Instead, the answer is a text paragraph. For English LFQA, we utilize the MultiMedQA dataset from \cite{singhal2023towards}, which contains 1066 questions curated from the HealthSearchQA \cite{singhal2023large}, LiveQA \cite{abacha2017overview}, MedicationQA \cite{abacha2019bridging} datasets. For Chinese LFQA, we collect 1,253 user queries from an online medical consultation platform\footnote{The name of the online medical consultation platform will be revealed upon acceptance.}. An expert panel ensures the safety of the dataset. This dataset is referred to as MRAG-CLFQA.

\textbf{Information extraction (IE).} To evaluate how LLMs powered by the RAG mechanism can perform in the medical information extraction tasks, we consider the following three tasks: (a) DDI \cite{herrero2013ddi}, which asks a model to extract triplets that reflects how drugs interact with one another. (b) ChemProt \cite{taboureau2010chemprot}, extracting the relationships among diseases, drugs, and genes from medical articles. (c) CMeIE \cite{guan2020cmeie}, which asks a model to extract triplets of 43 different relation types. The first two tasks are in English, and the last in Chinese. 


\textbf{Link prediction (LP).} The link prediction task \cite{kumar2020link} is suitable for evaluating LLMs since it directly checks whether LLMs correctly grasp the world knowledge and is of great importance for applications like drug repurposing \cite{aruna2022survey}. In MRAG, we consider the following two tasks: (a) ADInt \cite{xiao2024repurposing}, which is a dataset for identifying new pharmaceutical interventions (PI) for Alzheimer’s Disease (AD). Our MRAG-Bench randomly selects 1,500 samples from the ADInt testing set. (b) DRKG \cite{ioannidis2020drkg} is a knowledge graph investigating graph-based drug repurposing algorithms for COVID-19. We randomly select 1,500 drug-disease triplets. We reformulate ADInt and DRKG as multiple-choice QA tasks in which two medical entities are given in the prompt, and the model needs to determine the relation types.


In summary, MRAG investigates how LLM RAG systems perform on four cohorts of tasks, 13 different datasets, and a total of 14,816 test samples.

\subsection{Evaluation metrics}
\label{subsec:eval_metrics}

We use objective metrics can evaluate models on the  MCQA, IE, and LP task cohorts. For the LFQA tasks, since there are no standard answers, we conducted a series of model- and human-based evaluations.

\begin{figure*}
\centering
\includegraphics[width=0.78\textwidth]{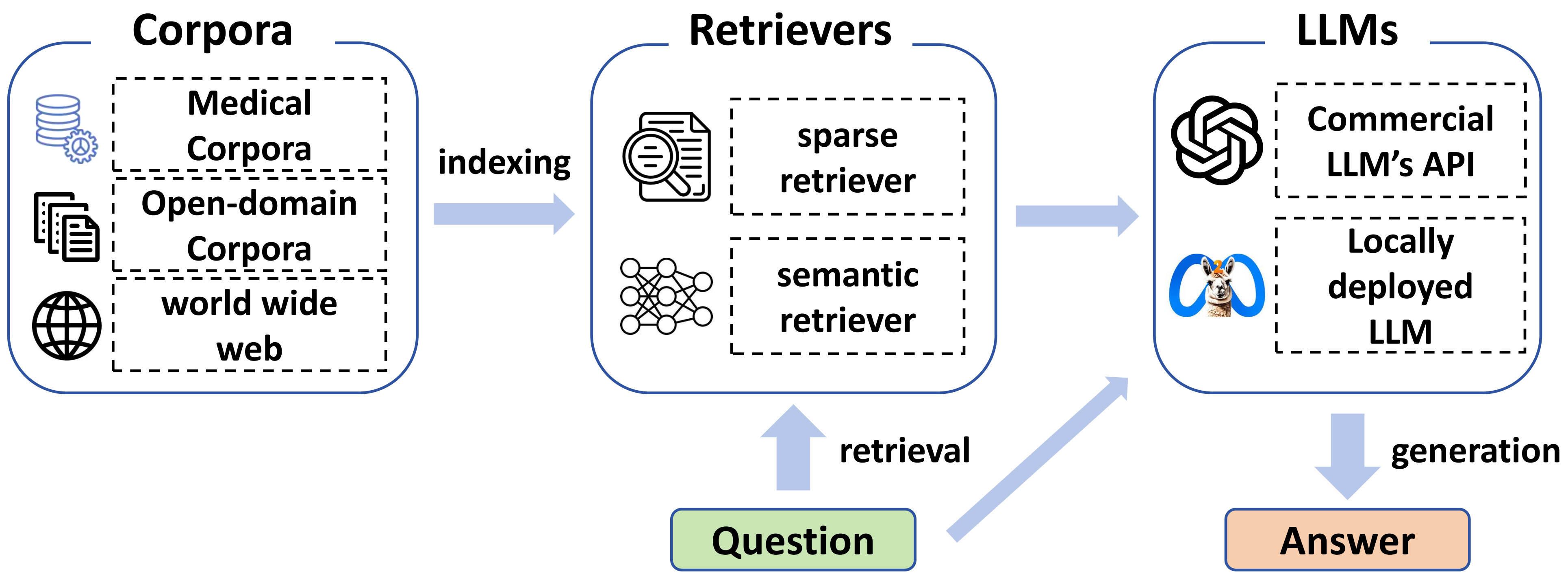} 
\caption{Framework of the MRAG toolkit, demonstrating each of its components. }
\label{fig:architecture}
\vspace{-8pt}
\end{figure*}

\section{RAG system }
\label{sec:rag_system}

To comprehensively evaluate how different LLM-based RAG systems perform on our MRAG benchmark, we propose MRAG-Toolkit, a toolkit with systematic implementations of RAG for medical QA. As shown in Figure \ref{fig:architecture}, the MRAG-Toolkit consists of the following major components: (a) the Corpus from which the supporting documents are retrieved; (b) Retriever, the model or method for coducting efficient retrieval on the retrieval corpus; (c) Response generator, the LLM used to summarize the supporting document and generate the final response; (d) Prompting strategies, the strategies that instruct the response generator how to summarize, reason, reflect, and organize the final response.

\section{Results}
\label{sec:experiment_results}

We systematically evaluate LLMs on our MRAG-Bench benchmark, providing a multi-dimensional analysis of different components in RAG for medicine.

\subsection{Results on the closed-form tasks}

We first benchmark various LLMs on the MCQA, IE, and LP tasks in the MRAG bench, both w/o. RAG and w. RAG. The COT-Refine strategy is used to elicit responses. All the LLMs utilize the nucleus sampling strategy for decoding. The temperature parameter is set to 0.7, the top\_p is set to 0.8, and the repetition penalty is set to 1.05. The combined corpus is used as the source of referential documents. The retriever is BGE-base, and eight snippets are retrieved for each query text and concatenated to the prompt if using RAG.


As shown in Table \ref{tab:mcqa_results}, without RAG, GPT-4 significantly outperforms other competitors, with an average score of 80.2\% on the MCQA cohort and 10.5\% on the IE cohort, 47.7\% on the LP cohort. Even the strongest open-sourced LLMs, Mixtral-8x22B,  can only achieve about 74.2\% of the GPT-4's performance on the MCQA tasks. In comparison, RAG helps all the models improve significantly on the MRAG-Bench regarding the average scores on the three cohorts. These results suggest RAG's great potential to enhance LLMs' zero-shot capability to answer users' medical questions or conduct medical knowledge discovery and make the LLMs more reliable. In addition, the following observations can be made: (a) Without RAG, MEDITRON, the domain-specific LLM, outperforms the open-domain LLM, Qwen2.5-72B, on the MCQA tasks. However, with the help of MRAG, Qwen2.5-72B outperforms MEDITRON on all three MRAG task cohorts. The intuition is that Qwen2.5-72B is better at following the instructions in MRAG prompts and incorporating the information retrieved from the documents in the reasoning steps. (b) MRAG is not beneficial on all of the MRAG tasks. MRAG negatively impacts the MMNLU-Med task for most of the LLMs evaluated. And relatively smaller LLMs, LlaMA-3, MEDITRON, and PMC-LlaMA, can not utilize the retrieved documents to improve the performance on MedQA. The results show that the retrieved documents may distract an LLM if it pays too much attention to the wrong document.

\begin{table*}
\centering
\resizebox{0.86\textwidth}{!}{
\renewcommand\arraystretch{1.05}
\begin{tabular}{cccccccc}
\hline
\multirow{2}*{\textbf{Method}}  &  \multirow{2}*{\makecell[c]{\textbf{Prompting}  \\ \textbf{method}} }  &   \multicolumn{5}{c}{ \textbf{MCQA tasks in MRAG-Bench} }   &  \multirow{2}*{\textbf{Avg.}}  \\ 
&      &      MMLU-Med   &   MedQA  &   MedMCQA  &   PubMedQA   &  BioASQ  &   \\
\hline

\hline

\multirow{2}*{\makecell[c]{\textbf{GPT-4}  \\ (gpt-4o)} }   & w/o. RAG   &  \textbf{93.1}   \textcolor{lightgray}{$\pm$ 1.2}     &    86.5  \textcolor{lightgray}{$\pm$ 0.9}     &   \textbf{76.2}  \textcolor{lightgray}{$\pm$ 0.8}   &   57.2  \textcolor{lightgray}{$\pm$ 1.4}    &        88.1  \textcolor{lightgray}{$\pm$ 1.1}   &   80.2  \\

&  w. RAG    &   92.5 \textcolor{lightgray}{$\pm$ 0.9} &     \textbf{89.3}  \textcolor{lightgray}{$\pm$ 1.0}   &  75.8 
 \textcolor{lightgray}{$\pm$ 0.7}   &   \textbf{78.3} \textcolor{lightgray}{$\pm$ 1.2}   &    \textbf{90.5}  \textcolor{lightgray}{$\pm$ 0.9}  &   \textbf{85.2}  \\
\hdashline

\multirow{2}*{\makecell[c]{\textbf{GPT-3.5}  \\ (gpt-3.5-turbo)} }   
 &  w/o. RAG   &   78.9 \textcolor{lightgray}{$\pm$ 1.3}   &   61.5 \textcolor{lightgray}{$\pm$ 0.8}  &   56.2 \textcolor{lightgray}{$\pm$ 1.1}   
&    36.7  \textcolor{lightgray}{$\pm$ 1.9} 
 &   76.9 \textcolor{lightgray}{$\pm$ 1.0}    &   62.0  \\
 
&  w. RAG   &   78.2 \textcolor{lightgray}{$\pm$ 1.0}  &  67.3 \textcolor{lightgray}{$\pm$ 0.7} &   57.8 \textcolor{lightgray}{$\pm$ 0.9}   &    68.7 \textcolor{lightgray}{$\pm$ 1.7}   &   86.4 \textcolor{lightgray}{$\pm$ 0.8}   &  71.7 \\
\hdashline


\multirow{2}*{\makecell[c]{\textbf{Tongyi Qwen}  \\ (qwen\_max)} } &  w/o. RAG   &  77.7  \textcolor{lightgray}{$\pm$ 1.1}  &  62.6   \textcolor{lightgray}{$\pm$ 1.1}  &  59.7  \textcolor{lightgray}{$\pm$ 0.9}  &  37.1 \textcolor{lightgray}{$\pm$ 1.6}  &  76.3 \textcolor{lightgray}{$\pm$ 1.2}   &   62.7   \\
&  w. RAG   &   77.1 \textcolor{lightgray}{$\pm$ 1.2}  &  66.7 \textcolor{lightgray}{$\pm$ 0.9 }  &  60.2 \textcolor{lightgray}{$\pm$ 0.8}  &  75.3 \textcolor{lightgray}{$\pm$ 1.8} &  88.1  \textcolor{lightgray}{$\pm$ 1.3}   &    73.5       \\
\hdashline


\multirow{2}*{\makecell[c]{\textbf{Mixtral}  \\ (8 * 22B)} } &  w/o. RAG   &   75.5 \textcolor{lightgray}{$\pm$ 1.5}  &  59.3  \textcolor{lightgray}{$\pm$ 1.2}   &     53.4  \textcolor{lightgray}{$\pm$ 0.7}     &   35.8 \textcolor{lightgray}{$\pm$ 1.9}      &  73.8 \textcolor{lightgray}{$\pm$ 1.4}    &  59.5    \\
&  w. RAG    &   75.1 \textcolor{lightgray}{$\pm$ 1.4}  &   62.5  \textcolor{lightgray}{$\pm$ 1.1}  &  54.0  \textcolor{lightgray}{$\pm$ 0.8}  &  74.1 \textcolor{lightgray}{$\pm$ 1.7}  &   84.2 \textcolor{lightgray}{$\pm$ 1.2}   &   69.9  \\
\hdashline

\multirow{2}*{\makecell[c]{\textbf{LlaMA-3}  \\ (70B)} } &  w/o. RAG     &   61.6 \textcolor{lightgray}{$\pm$ 1.2}  &  54.1 \textcolor{lightgray}{$\pm$ 1.1}   &     44.5 \textcolor{lightgray}{$\pm$ 0.9}     &   32.6  \textcolor{lightgray}{$\pm$ 1.7}     &  61.6  \textcolor{lightgray}{$\pm$ 1.7}    &   50.9   \\
&  w. RAG       &    61.4   \textcolor{lightgray}{$\pm$ 1.2}   &  53.6  \textcolor{lightgray}{$\pm$ 1.0}  &   45.2 \textcolor{lightgray}{$\pm$ 0.7}   &  
  63.5 \textcolor{lightgray}{$\pm$ 1.5}  &     70.2  \textcolor{lightgray}{$\pm$ 1.6}    &    58.8  \\
\hdashline

\multirow{2}*{\makecell[c]{\textbf{Qwen2.5}  \\ (72B)} } &  w/o. RAG   &  70.6 \textcolor{lightgray}{$\pm$ 0.9}  &  55.6 \textcolor{lightgray}{$\pm$ 0.9}  &  43.9 \textcolor{lightgray}{$\pm$ 0.8}   &   30.9 \textcolor{lightgray}{$\pm$ 1.6}  &  64.3 \textcolor{lightgray}{$\pm$ 1.6}    &  53.1  \\
&  w. RAG    &  68.5 \textcolor{lightgray}{$\pm$ 1.0} &  56.9 \textcolor{lightgray}{$\pm$ 0.7}  &  43.0 \textcolor{lightgray}{$\pm$ 0.8}     &   69.2 \textcolor{lightgray}{$\pm$ 1.9}  &  81.7 \textcolor{lightgray}{$\pm$ 1.6}    &   63.9   \\
\hdashline

\multirow{2}*{\makecell[c]{\textbf{MEDITRON}  \\ (70B)} } &  w/o. RAG   &   64.9  \textcolor{lightgray}{$\pm$ 1.6}  &  51.6 \textcolor{lightgray}{$\pm$ 1.1}  &   46.7  \textcolor{lightgray}{$\pm$ 1.0}  &   43.4 \textcolor{lightgray}{$\pm$ 1.7}  &   68.4 \textcolor{lightgray}{$\pm$ 1.9}    &    55.0  \\
&  w. RAG    &  65.4   \textcolor{lightgray}{$\pm$ 1.5}  &   49.5 \textcolor{lightgray}{$\pm$ 1.1}  &   45.9  \textcolor{lightgray}{$\pm$ 0.9}  &   53.4 \textcolor{lightgray}{$\pm$ 1.6}  &   76.8  \textcolor{lightgray}{$\pm$ 1.6}   &    58.2   \\
\hdashline

\multirow{2}*{\makecell[c]{\textbf{PMC-LlaMA}  \\ (13B)} } &  w/o. RAG   &    52.2  \textcolor{lightgray}{$\pm$ 1.7}  &   44.3  \textcolor{lightgray}{$\pm$ 1.2}  &   46.5 \textcolor{lightgray}{$\pm$ 1.1}  &   35.8 \textcolor{lightgray}{$\pm$ 2.2}  &   63.1 \textcolor{lightgray}{$\pm$ 1.7}    &  48.4  \\
&  w. RAG    &   52.5  \textcolor{lightgray}{$\pm$ 1.4}  &   42.6 \textcolor{lightgray}{$\pm$ 1.3}  &   48.3 \textcolor{lightgray}{$\pm$ 1.0}  &   54.0 \textcolor{lightgray}{$\pm$ 2.1}   &    65.2 \textcolor{lightgray}{$\pm$ 1.5}   &  52.5   \\

\hline
\end{tabular}}
\caption{Benchmark results of different backbone LLMs on the multi-choice QA tasks in MRAG-Bench. We report the average accuracy in percentages on five different runs, along with the standard deviations in the light-grey color. }
\vspace{-6pt}
\label{tab:mcqa_results}
\end{table*}

\subsection{Results on the LFQA tasks}

For the LFQA task, we first ask GPT-4 and GPT-3.5 to generate responses, with or without RAG. Then, we put these four models into an arena where each match randomly selects a pair of responses for the same query. They are judged by GPT-4 to determine which model's response wins, loses, or is a draw with the other in terms of the evaluation axes described in Section \ref{subsec:eval_metrics}. These matches will continue until all the model pairs have at least 80 matches for each evaluation axis. In this work, we will ask GPT-4 to serve as an unbiased judge \cite{zheng2024judging,zheng2024sca,zhang2024milora} and ask medical experts to annotate a part of the matches to ensure the quality.

After the models conduct matches in the arena, we use the Elo rating system to rank the models along the four axes. The Elo rating system \cite{glickman2010uscf} is a method for calculating the relative skill levels of players in competitive games. In this work, we set the initial score of each LLM as 1000 and the $ K$ factor to 40 in the arena. The Elo scores are presented in Table \ref{tab:elo_results}.   

From Table \ref{tab:elo_results}, we can see that RAG can effectively improve the LLM's \emph{usefulness}, \emph{knowledge}, and \emph{reasoning} in the LFQA task. However, the \emph{readability} of LLMs declines with RAG. The intuition is that the LLMs tend to answer more formally verbatim with RAG, making it slightly more difficult for the layer-persons to read.

\begin{table*}
\centering
\resizebox{0.80\textwidth}{!}{
\renewcommand\arraystretch{1.2}
\begin{tabular}{cccccc}
\hline
\textbf{Model}  &  \textbf{Prompting method}  &   \textbf{Usefulness}    &   \textbf{Readability}    &   
 \textbf{Knowledge}    &    \textbf{Reasoning}     \\ 
\hline

\multirow{2}*{\makecell[c]{\textbf{GPT-4}  \\ (gpt-4o)} }   & w/o. RAG    &  931.7    &   \textbf{1179.5}   &  1062.4   &    990.9     \\
&  w. RAG     &  \textbf{1270.9}    &     1082.4   &     696.1   &     \textbf{1259.4}     \\
\hdashline

\multirow{2}*{\makecell[c]{\textbf{GPT-3.5}  \\ (gpt-3.5-turbo)} } & w/o. RAG    &   811.7   &  873.4    &   \textbf{1345.8}   &       773.5       \\
&  w. RAG  &   985.5    &   864.6   &   895.6    &       976.0    \\

\hline
\end{tabular}}
\caption{Results of different backbone LLMs on the LFQA task in MRAG-Bench. The Elo rating scores on each evaluation axis are reported. The highest scores are in bold.   }
\vspace{-2pt}
\label{tab:elo_results}
\end{table*}

\subsection{Ablation studies}

\begin{table}[tb!]
\centering
\resizebox{0.49\textwidth}{!}{
\begin{tabular}{cccc}
\hline
\textbf{Retriever}  &   \textbf{PubMedQA}  &     \textbf{MedQA}     &    \textbf{DRKG}  \\
\hline

\hline

BGE-base   &     68.7 \textcolor{lightgray}{$\pm$ 1.7}    &   67.3 \textcolor{lightgray}{$\pm$ 0.7}       &     \textbf{35.4} \textcolor{lightgray}{$\pm$ 1.7}      \\

BM25   &   61.3  \textcolor{lightgray}{$\pm$ 2.1}    &    62.1  \textcolor{lightgray}{$\pm$ 0.9}   &     31.2   \textcolor{lightgray}{$\pm$ 1.4}   \\

MedCPT   &    67.9  \textcolor{lightgray}{$\pm$ 1.6}   &  65.8  \textcolor{lightgray}{$\pm$ 1.1}    &   34.4  \textcolor{lightgray}{$\pm$ 1.5}     \\

E5-Mistral-7B   &   68.6  \textcolor{lightgray}{$\pm$ 1.9}   &    67.4  \textcolor{lightgray}{$\pm$ 0.7}     &  35.3   \textcolor{lightgray}{$\pm$ 1.8}   \\

RRF   &   \textbf{68.9}  \textcolor{lightgray}{$\pm$ 1.6}    &   \textbf{67.5}  \textcolor{lightgray}{$\pm$ 0.8}   &   35.1  \textcolor{lightgray}{$\pm$ 1.6}   \\

\hline
\end{tabular}}
\caption{Comparison of different retrievers for MRAG. The LLM is GPT-3.5.  }
\vspace{-4pt}
\label{tab:ablation_retrievers}
\end{table}

\textbf{Comparison of different retrievers} \quad In this subsection, we now investigate how different retrievers affect the performances of RAG systems. Table \ref{tab:ablation_retrievers} reports the performance of GPT-3.5 with RAG on PubmedQA, MedQA, and DRKG with the help of different retrievers. The domain-specific retrievers, MedCPT, perform slightly worse than BGE-base. The heavier retriever, E5-mistral-7B, performs comparably to the BGE base. The results show that: (a) by large-scale contrastive learning, BGE-base is also effective in retrieving domain-specific documents. (b) domain-specific pretraining does not provide clear advantages against the open-domain retrievers. Table \ref{tab:ablation_retrievers} also demonstrates that RRF, the fusion of BGE-base and MedCPT, outperforms the two component retrievers on two of the three tasks. However, employing RRF in applications means multiple retrievers have to be deployed.

\begin{table}[tb!]
\centering
\resizebox{0.49\textwidth}{!}{
\begin{tabular}{cccc}
\hline
\textbf{Corpus}  &   \textbf{PubMedQA}  &     \textbf{MedQA}     &    \textbf{DRKG}  \\
\hline

\hline

Combined corpus   &     68.7 \textcolor{lightgray}{$\pm$ 1.7}    &   67.3 \textcolor{lightgray}{$\pm$ 0.7}       &     35.4 \textcolor{lightgray}{$\pm$ 1.7}      \\

Medical corpus   &   68.5  \textcolor{lightgray}{$\pm$ 1.9}    &    67.0  \textcolor{lightgray}{$\pm$ 1.1}   &     35.5   \textcolor{lightgray}{$\pm$ 1.3}   \\

Open-domain corpus   &    57.4  \textcolor{lightgray}{$\pm$ 1.9}   &  63.6  \textcolor{lightgray}{$\pm$ 1.5}    &   31.9  \textcolor{lightgray}{$\pm$ 1.3}     \\

World Wide Web   &   60.2  \textcolor{lightgray}{$\pm$ 2.1}   &    66.3 \textcolor{lightgray}{$\pm$ 1.1}     &  33.8   \textcolor{lightgray}{$\pm$ 1.8}   \\

\hline
\end{tabular}}
\caption{Comparison of different corpora for MRAG. The LLM is GPT-3.5.  }
\vspace{-8pt}
\label{tab:ablation_corpus}
\end{table}

\textbf{Effects of different corpora} \quad Table \ref{tab:ablation_corpus} reports the performance of GPT-3.5 w. RAG on PubmedQA, MedQA, and DRKG, with different corpus for document retrieval. The results show that: (a) The performance of LLMs on the MRAG tasks is highly related to the referential corpus. For example, the open-domain corpus is significantly less helpful than the medical corpus for the PubMedQA task. (b) A simple combination of the two corpora of different domains does not affect the average accuracy on the three medical tasks. (c) The World Wide Web is helpful for the MedQA task but is less beneficial for the other two tasks since these two tasks rely on the medical literature.

\begin{table}[tb!]
\centering
\resizebox{0.49\textwidth}{!}{
\begin{tabular}{cccc}
\hline
\textbf{Prompting}  &   \textbf{PubMedQA}  &     \textbf{MedQA}     &    \textbf{DRKG}  \\
\hline

\hline

COT-Refine   &     68.7 \textcolor{lightgray}{$\pm$ 1.7}    &   67.3 \textcolor{lightgray}{$\pm$ 0.7}       &     35.4 \textcolor{lightgray}{$\pm$ 1.7}      \\

Chain-of-thought &   67.3  \textcolor{lightgray}{$\pm$ 1.6}    &    66.3  \textcolor{lightgray}{$\pm$ 0.9}   &     35.2   \textcolor{lightgray}{$\pm$ 1.6}   \\

Direct answer  &    63.1  \textcolor{lightgray}{$\pm$ 2.1}   &  64.5  \textcolor{lightgray}{$\pm$ 1.3}    &   30.8  \textcolor{lightgray}{$\pm$ 1.8}     \\

\hline
\end{tabular}}
\caption{Comparison of different prompting strategies for eliciting responses. The LLM is GPT-3.5.  }
\vspace{-2pt}
\label{tab:ablation_prompting}
\end{table}

\textbf{Comparison of different prompting strategies} \quad Table \ref{tab:ablation_prompting} reports the performance of GPT-3.5 w. RAG on PubmedQA, MedQA, and DRKG, with prompting strategies for eliciting responses. The results show that: (a) compared with the chain-of-thought strategy and direct answer strategy, the COT-Refine improves the LLM's accuracy by reflecting on its previous answer and improving by better incorporating the referential documents and changing the reasoning steps. (b) The direct answer strategy results show that directly answering a question without reasoning steps leads to performance degradation.

\begin{figure}
    \centering
\subfigure{%
\includegraphics[width=0.235\textwidth]{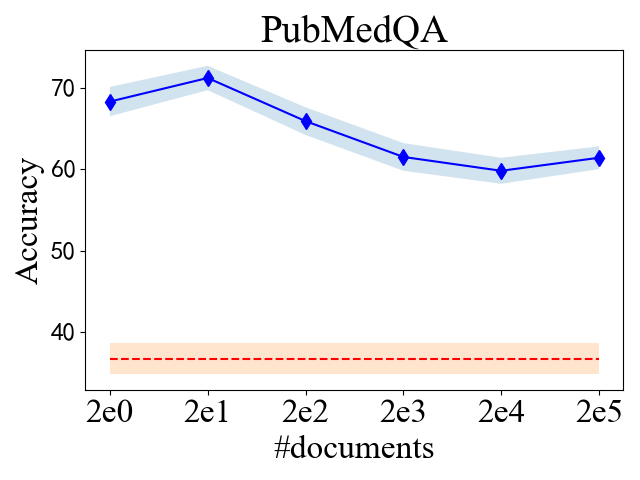}
\label{subfig:PubMedQA_different_topk}
}\hspace{-8pt}
\subfigure{%
\includegraphics[width=0.235\textwidth]{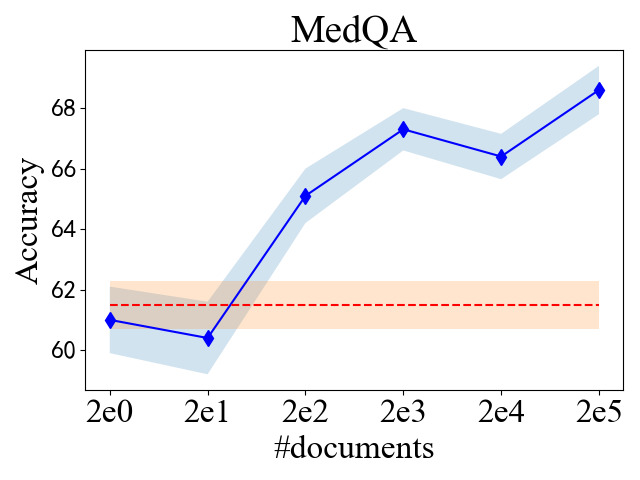}
\label{subfig:MedQA_different_topk}
}
\caption{Effects of \#documents retrieved.}
\label{fig:different_topk}
\end{figure}

\textbf{Investigating the scaling laws in RAG} \quad We first explore how the performance of MRAG scales with the increase in the number of documents retrieved and concatenated for medical QA tasks. In these experiments, we use GPT-3.5 as the backbone LLM and utilize the RRF retriever and the combined corpus for retrieval. Figure \ref{fig:different_topk} shows the scaling curves of MRAG on the PubMedQA and MedQA tasks with different numbers of snippets $k \in \{1, 2, 4, ..., 64\}$. The scaling curves are quite different for different tasks. On the MedQA task, we see roughly log-linear curves in the scaling plot for $k \leq 32$. However, on the PubMedQA task, the ground truth documents can be accurately retrieved, and MRAG presents higher performance when $k \leq 2$. Moreover, with $k$ increases, more irrelevant documents are included in the prompt, hurting the accuracy.

\begin{figure}
    \centering
\subfigure{%
\includegraphics[width=0.235\textwidth]{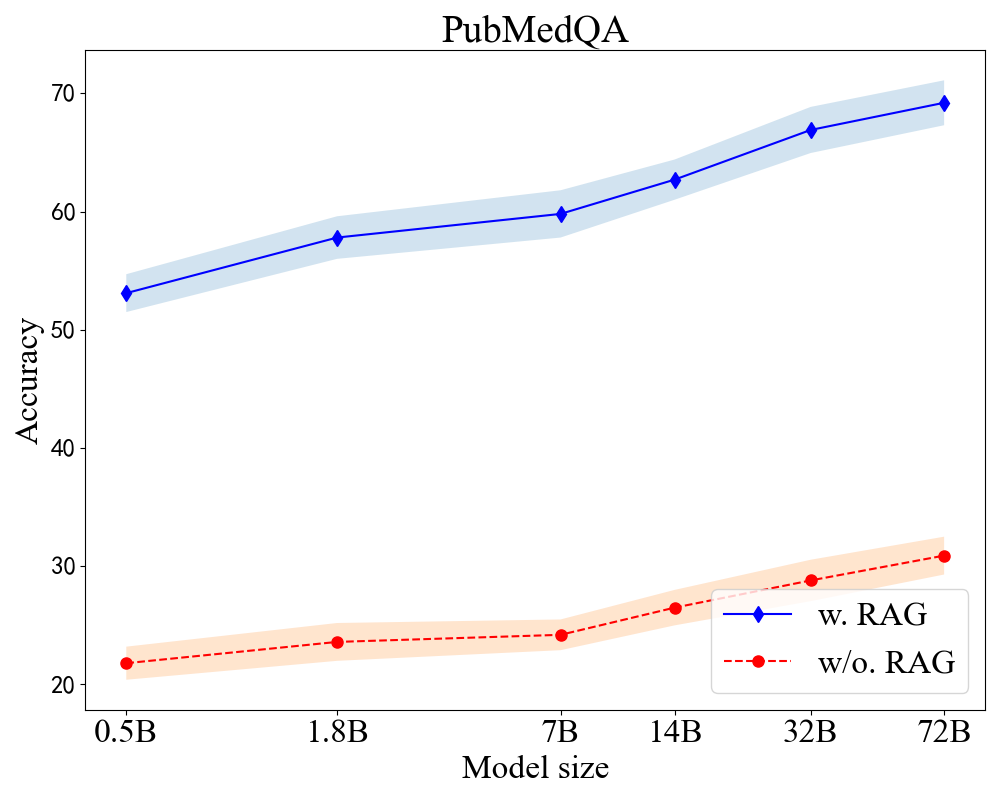}
\label{subfig:PubMedQA_different_model_size}
}\hspace{-8pt}
\subfigure{%
\includegraphics[width=0.235\textwidth]{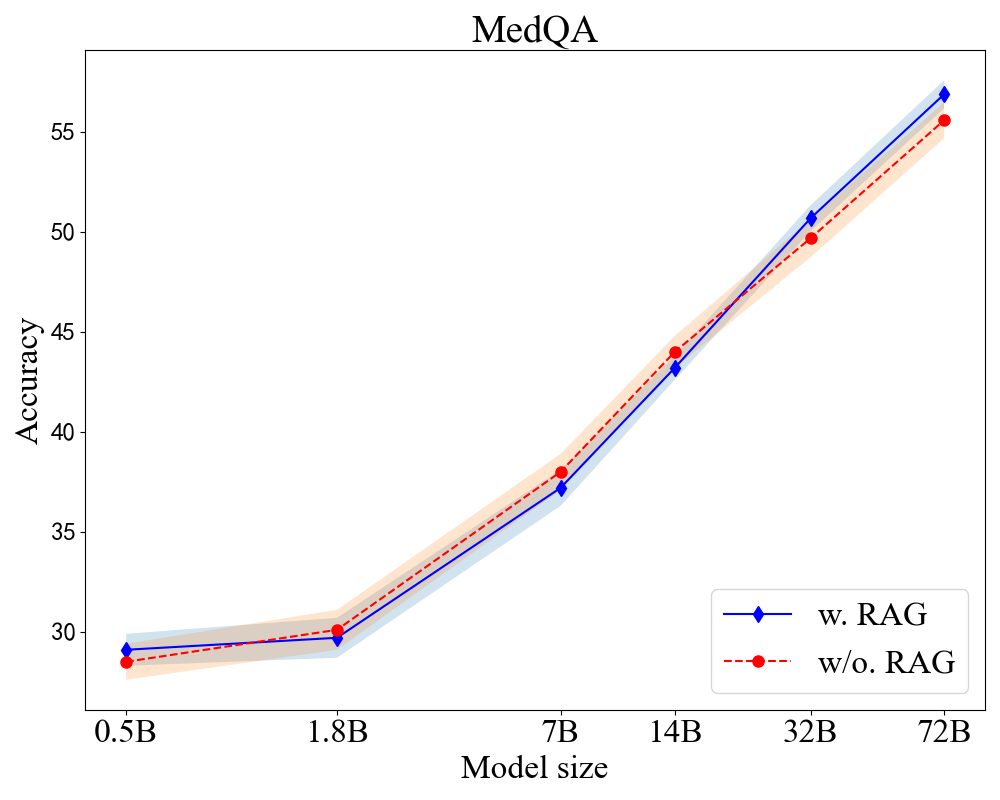}
\label{subfig:MedQA_different_model_size}
}
\caption{The scaling laws of LLMs on the MRAG tasks, with or without RAG.}
\label{fig:different_llm_sizes}
\end{figure}

To investigate the scaling law of model sizes, we use the Qwen2.5 models of different sizes (0.5B, 1.8B, 7B, 14B, 32B, 72B) while keeping the other settings fixed. Figure \ref{fig:different_llm_sizes} shows the scaling curves of Qwen model sizes on the PubMedQA and MedQA tasks, with or without RAG. The four curves are roughly log-linear, demonstrating that the scaling law of LLMs \cite{kaplan2020scaling} applies under RAG. In addition, the performance gaps between the RAG and non-RAG curves increase slightly as the model scales up since larger models are better at incorporating the referential documents into the reasoning steps.

\section{Conclusions}
\label{sec:conclusions}

In this work, we presented the Medical Retrieval-Augmented Generation benchmark (MRAG-Bench) and the MRAG-Toolkit, designed to systematically evaluate and enhance the performance of LLMs through Retrieval-Augmented Generation (RAG). Our MRAG-Bench spans four task cohorts across English and Chinese, providing a robust evaluation framework for LLM-based RAG systems. The MRAG-Toolkit supports various retrieval approaches, algorithms, and LLMs, enabling a detailed investigation of how different components influence performance. Our experiments demonstrated that RAG significantly improves LLM reliability in all MRAG tasks. We observed that the choice of the referential corpus, retrieval methods, and prompting strategies heavily influence LLM performance. Additionally, larger LLMs benefit more from RAG, although there is a trade-off in readability for long-form question answering. The MRAG-Bench and MRAG-Toolkit will serve as valuable resources for the research community, fostering further advancements in RAG and its applications in the medical industry.

\section*{Limitations} 
\label{sec:limitations}

Despite the fact that we provide extensive experiments of the MRAG benchmark in this work, the following limitations remain: (a) Powerful language models like Gemini \cite{reid2024gemini}, Claude-3\footnote{\url{https://www.anthropic.com/news/claude-3-family}.}, Grok\footnote{https://github.com/xai-org/grok-1} are not evaluated due to resource limitation. (b) There are literature in RAG that adopt more complicated workflow than our MRAG system (in Figure \ref{fig:architecture}), such as iterative retrieval \cite{zhang2023repocoder,jiang2023active}. These more advanced RAG strategies have not been evaluated in our current version, but we will address this aspect in our updated version.

\section*{Ethical considerations}
\label{sec:broader_impacts}

The advancement of Large Language Models (LLMs) and their integration with Retrieval-Augmented Generation (RAG) systems have significant implications for various domains, particularly in high-stakes fields like healthcare. The Medical Retrieval-Augmented Generation (MRAG) benchmark and MRAG-Toolkit developed in this study aim to enhance the reliability and accuracy of LLMs in medical question answering (QA). Our work leads to the following positive or negative sociatal impacts:
\begin{itemize}
    \item Positive Societal Impacts:

    \begin{itemize}
        \item Enhanced Medical Information Access: By integrating Retrieval-Augmented Generation (RAG) with Large Language Models (LLMs), our work can significantly improve access to up-to-date and reliable medical information. This is particularly valuable in clinical settings where timely and accurate information is crucial for patient care. The MRAG system can assist healthcare professionals in making more informed decisions, potentially leading to better patient outcomes.

        \item Transparency and Accountability: The RAG approach enhances the transparency of LLMs by grounding their responses in retrieved documents. This can foster trust in AI systems as users can trace back the source of the information provided. Such transparency is essential in the medical field where the provenance of information can impact clinical decisions.

        \item Open-Sourced Toolkit: The MRAG-Toolkit we have developed is open-sourced, promoting collaboration and further research in the field. By providing a standardized evaluation framework, we enable other researchers to build upon our work, accelerating advancements in medical AI and contributing to the broader scientific community.

    \end{itemize}

    \item Negative Societal Impacts:

    \begin{itemize}
        \item Reliance on AI Systems: While RAG enhances the reliability of LLMs, there is a risk that over-reliance on AI systems might emerge, potentially leading to complacency among healthcare professionals. It is crucial to maintain a balance where AI serves as a supportive tool rather than a replacement for professional judgment.

        \item Impact on Healthcare Workforce: The introduction of advanced AI systems like MRAG may impact the job market for certain roles within the healthcare sector. While AI can augment human capabilities, it may also lead to job displacement, necessitating a focus on retraining and upskilling affected workers.

        \item Potential for Bias: The retrieved documents and underlying datasets may contain biases that could be propagated or even amplified by the MRAG system. Ensuring that the sources used for retrieval are diverse and unbiased is essential to mitigate this risk. It is our duty to further study the bias issue of MRAG-Bench. 
        
    \end{itemize}
    
\end{itemize}

By carefully considering these positive and negative impacts, our work aims to contribute to the development and deployment of responsible LLM-based technologies in the medical domain.

\bibliography{custom}

@inproceedings{sun-etal-2022-simple,
    title = "A Simple Hash-Based Early Exiting Approach For Language Understanding and Generation",
    author = "Sun, Tianxiang  and
      Liu, Xiangyang  and
      Zhu, Wei  and
      Geng, Zhichao  and
      Wu, Lingling  and
      He, Yilong  and
      Ni, Yuan  and
      Xie, Guotong  and
      Huang, Xuanjing  and
      Qiu, Xipeng",
    booktitle = "Findings of the Association for Computational Linguistics: ACL 2022",
    month = may,
    year = "2022",
    address = "Dublin, Ireland",
    publisher = "Association for Computational Linguistics",
    url = "https://aclanthology.org/2022.findings-acl.189",
    doi = "10.18653/v1/2022.findings-acl.189",
    pages = "2409--2421",
}

@article{Zhang2021AutomaticSN,
  title={Automatic Student Network Search for Knowledge Distillation},
  author={Zhexi Zhang and Wei Zhu and Junchi Yan and Peng Gao and Guowang Xie},
  journal={2020 25th International Conference on Pattern Recognition (ICPR)},
  year={2021},
  pages={2446-2453}
}

@inproceedings{zhu-etal-2021-gaml,
	title = "{GAML}-{BERT}: Improving {BERT} Early Exiting by Gradient Aligned Mutual Learning",
	author = "Zhu, Wei  and
	Wang, Xiaoling  and
	Ni, Yuan  and
	Xie, Guotong",
	booktitle = "Proceedings of the 2021 Conference on Empirical Methods in Natural Language Processing",
	month = nov,
	year = "2021",
	address = "Online and Punta Cana, Dominican Republic",
	publisher = "Association for Computational Linguistics",
	url = "https://aclanthology.org/2021.emnlp-main.242",
	pages = "3033--3044",
}

@inproceedings{zhang-etal-2022-pcee,
    title = "{PCEE}-{BERT}: Accelerating {BERT} Inference via Patient and Confident Early Exiting",
    author = "Zhang, Zhen  and
      Zhu, Wei  and
      Zhang, Jinfan  and
      Wang, Peng  and
      Jin, Rize  and
      Chung, Tae-Sun",
    booktitle = "Findings of the Association for Computational Linguistics: NAACL 2022",
    month = jul,
    year = "2022",
    address = "Seattle, United States",
    publisher = "Association for Computational Linguistics",
    url = "https://aclanthology.org/2022.findings-naacl.25",
    doi = "10.18653/v1/2022.findings-naacl.25",
    pages = "327--338",
}

@article{Ding2022DeltaTA,
  title={Delta Tuning: A Comprehensive Study of Parameter Efficient Methods for Pre-trained Language Models},
  author={Ning Ding and Yujia Qin and Guang Yang and Fu Wei and Zonghan Yang and Yusheng Su and Shengding Hu and Yulin Chen and Chi-Min Chan and Weize Chen and Jing Yi and Weilin Zhao and Xiaozhi Wang and Zhiyuan Liu and Haitao Zheng and Jianfei Chen and Yang Liu and Jie Tang and Juan Li and Maosong Sun},
  journal={ArXiv},
  year={2022},
  volume={abs/2203.06904}
}

@inproceedings{li-etal-2019-pingan,
    title = "Pingan Smart Health and {SJTU} at {COIN} - Shared Task: utilizing Pre-trained Language Models and Common-sense Knowledge in Machine Reading Tasks",
    author = "Li, Xiepeng  and
      Zhang, Zhexi  and
      Zhu, Wei  and
      Li, Zheng  and
      Ni, Yuan  and
      Gao, Peng  and
      Yan, Junchi  and
      Xie, Guotong",
    booktitle = "Proceedings of the First Workshop on Commonsense Inference in Natural Language Processing",
    month = nov,
    year = "2019",
    address = "Hong Kong, China",
    publisher = "Association for Computational Linguistics",
    url = "https://aclanthology.org/D19-6011",
    doi = "10.18653/v1/D19-6011",
    pages = "93--98",
}

@inproceedings{zhu-etal-2021-discovering,
    title = "Discovering Better Model Architectures for Medical Query Understanding",
    author = "Zhu, Wei  and
      Ni, Yuan  and
      Wang, Xiaoling  and
      Xie, Guotong",
    booktitle = "Proceedings of the 2021 Conference of the North American Chapter of the Association for Computational Linguistics: Human Language Technologies: Industry Papers",
    month = jun,
    year = "2021",
    address = "Online",
    publisher = "Association for Computational Linguistics",
    url = "https://aclanthology.org/2021.naacl-industry.29",
    doi = "10.18653/v1/2021.naacl-industry.29",
    pages = "230--237",
}

@inproceedings{zuo-etal-2022-continually,
    title = "Continually Detection, Rapidly React: Unseen Rumors Detection Based on Continual Prompt-Tuning",
    author = "Zuo, Yuhui  and
      Zhu, Wei  and
      Cai, Guoyong GUET",
    booktitle = "Proceedings of the 29th International Conference on Computational Linguistics",
    month = oct,
    year = "2022",
    address = "Gyeongju, Republic of Korea",
    publisher = "International Committee on Computational Linguistics",
    url = "https://aclanthology.org/2022.coling-1.268",
    pages = "3029--3041",
}

@inproceedings{guo-etal-2021-global,
    title = "Global Attention Decoder for {C}hinese Spelling Error Correction",
    author = "Guo, Zhao  and
      Ni, Yuan  and
      Wang, Keqiang  and
      Zhu, Wei  and
      Xie, Guotong",
    booktitle = "Findings of the Association for Computational Linguistics: ACL-IJCNLP 2021",
    month = aug,
    year = "2021",
    address = "Online",
    publisher = "Association for Computational Linguistics",
    url = "https://aclanthology.org/2021.findings-acl.122",
    doi = "10.18653/v1/2021.findings-acl.122",
    pages = "1419--1428",
}

@Article{info:doi/10.2196/17653,
author="Sun, Haixia
and Xiao, Jin
and Zhu, Wei
and He, Yilong
and Zhang, Sheng
and Xu, Xiaowei
and Hou, Li
and Li, Jiao
and Ni, Yuan
and Xie, Guotong",
title="Medical Knowledge Graph to Enhance Fraud, Waste, and Abuse Detection on Claim Data: Model Development and Performance Evaluation",
journal="JMIR Med Inform",
year="2020",
month="Jul",
day="23",
volume="8",
number="7",
pages="e17653",
issn="2291-9694",
doi="10.2196/17653",
url="http://medinform.jmir.org/2020/7/e17653/",
}

@inproceedings{gao2023f,
  title={F-PABEE: Flexible-patience-based Early Exiting for Single-label and Multi-label text Classification Tasks},
  author={Gao, Xiangxiang and Zhu, Wei and Gao, Jiasheng and Yin, Congrui},
  booktitle={ICASSP 2023-2023 IEEE International Conference on Acoustics, Speech and Signal Processing (ICASSP)},
  pages={1--5},
  year={2023},
  organization={IEEE}
}

@article{qin2023chatgpt,
  title={Is ChatGPT a general-purpose natural language processing task solver?},
  author={Qin, Chengwei and Zhang, Aston and Zhang, Zhuosheng and Chen, Jiaao and Yasunaga, Michihiro and Yang, Diyi},
  journal={arXiv preprint arXiv:2302.06476},
  year={2023}
}

@article{hendrycks2020measuring,
  title={Measuring massive multitask language understanding},
  author={Hendrycks, Dan and Burns, Collin and Basart, Steven and Zou, Andy and Mazeika, Mantas and Song, Dawn and Steinhardt, Jacob},
  journal={arXiv preprint arXiv:2009.03300},
  year={2020}
}

@article{huang2023c,
  title={C-eval: A multi-level multi-discipline chinese evaluation suite for foundation models},
  author={Huang, Yuzhen and Bai, Yuzhuo and Zhu, Zhihao and Zhang, Junlei and Zhang, Jinghan and Su, Tangjun and Liu, Junteng and Lv, Chuancheng and Zhang, Yikai and Lei, Jiayi and others},
  journal={arXiv preprint arXiv:2305.08322},
  year={2023}
}

@article{li2023cmmlu,
  title={CMMLU: Measuring massive multitask language understanding in Chinese},
  author={Li, Haonan and Zhang, Yixuan and Koto, Fajri and Yang, Yifei and Zhao, Hai and Gong, Yeyun and Duan, Nan and Baldwin, Timothy},
  journal={arXiv preprint arXiv:2306.09212},
  year={2023}
}

@article{singhal2023large,
  title={Large language models encode clinical knowledge},
  author={Singhal, Karan and Azizi, Shekoofeh and Tu, Tao and Mahdavi, S Sara and Wei, Jason and Chung, Hyung Won and Scales, Nathan and Tanwani, Ajay and Cole-Lewis, Heather and Pfohl, Stephen and others},
  journal={Nature},
  pages={1--9},
  year={2023},
  publisher={Nature Publishing Group UK London}
}

@article{singhal2023towards,
  title={Towards expert-level medical question answering with large language models},
  author={Singhal, Karan and Tu, Tao and Gottweis, Juraj and Sayres, Rory and Wulczyn, Ellery and Hou, Le and Clark, Kevin and Pfohl, Stephen and Cole-Lewis, Heather and Neal, Darlene and others},
  journal={arXiv preprint arXiv:2305.09617},
  year={2023}
}

@article{Li2023UnifiedDR,
  title={Unified Demonstration Retriever for In-Context Learning},
  author={Xiaonan Li and Kai Lv and Hang Yan and Tianya Lin and Wei Zhu and Yuan Ni and Guo Tong Xie and Xiaoling Wang and Xipeng Qiu},
  journal={ArXiv},
  year={2023},
  volume={abs/2305.04320},
  url={https://api.semanticscholar.org/CorpusID:258557751}
}

@article{Touvron2023Llama2O,
  title={Llama 2: Open Foundation and Fine-Tuned Chat Models},
  author={Hugo Touvron and Louis Martin and Kevin R. Stone and Peter Albert and Amjad Almahairi and Yasmine Babaei and Nikolay Bashlykov and Soumya Batra and Prajjwal Bhargava and Shruti Bhosale and Daniel M. Bikel and Lukas Blecher and Cristian Cant{\'o}n Ferrer and Moya Chen and Guillem Cucurull and David Esiobu and Jude Fernandes and Jeremy Fu and Wenyin Fu and Brian Fuller and Cynthia Gao and Vedanuj Goswami and Naman Goyal and Anthony S. Hartshorn and Saghar Hosseini and Rui Hou and Hakan Inan and Marcin Kardas and Viktor Kerkez and Madian Khabsa and Isabel M. Kloumann and A. V. Korenev and Punit Singh Koura and Marie-Anne Lachaux and Thibaut Lavril and Jenya Lee and Diana Liskovich and Yinghai Lu and Yuning Mao and Xavier Martinet and Todor Mihaylov and Pushkar Mishra and Igor Molybog and Yixin Nie and Andrew Poulton and Jeremy Reizenstein and Rashi Rungta and Kalyan Saladi and Alan Schelten and Ruan Silva and Eric Michael Smith and R. Subramanian and Xia Tan and Binh Tang and Ross Taylor and Adina Williams and Jian Xiang Kuan and Puxin Xu and Zhengxu Yan and Iliyan Zarov and Yuchen Zhang and Angela Fan and Melanie Kambadur and Sharan Narang and Aurelien Rodriguez and Robert Stojnic and Sergey Edunov and Thomas Scialom},
  journal={ArXiv},
  year={2023},
  volume={abs/2307.09288},
  url={https://api.semanticscholar.org/CorpusID:259950998}
}

@InProceedings{Text2dt,
author="Zhu, Wei
and Li, Wenfeng
and Wang, Xiaoling
and Ji, Wendi
and Wu, Yuanbin
and Chen, Jin
and Chen, Liang
and Tang, Buzhou",
editor="Tang, Buzhou
and Chen, Qingcai
and Lin, Hongfei
and Wu, Fei
and Liu, Lei
and Hao, Tianyong
and Wang, Yanshan
and Wang, Haitian
and Lei, Jianbo
and Li, Zuofeng
and Zong, Hui",
title="Extracting Decision Trees from Medical Texts: An Overview of the Text2DT Track in CHIP2022",
booktitle="Health Information Processing. Evaluation Track Papers",
year="2023",
publisher="Springer Nature Singapore",
address="Singapore",
pages="89--102",
isbn="978-981-99-4826-0"
}

@misc{yue2023-TCMEB,
      title={TCMEB: Performance Evaluation of Large Language Models Based on Traditional Chinese Medicine Benchmarks}, 
      author={Wenjing Yue, Wei Zhu and Xiaoling Wang},
      year={2023},
      publisher = {GitHub},
      journal = {GitHub repository},
      howpublished = {\url{https://github.com/ywjawmw/ShenNong-TCM-Evaluation-BenchMark}},
}

@inproceedings{Zhang2023NAGNERAU,
  title={NAG-NER: a Unified Non-Autoregressive Generation Framework for Various NER Tasks},
  author={Xinpeng Zhang and Ming Tan and Jingfan Zhang and Wei Zhu},
  booktitle={Annual Meeting of the Association for Computational Linguistics},
  year={2023},
  url={https://api.semanticscholar.org/CorpusID:259370837}
}

@InProceedings{text2dt_shared_task,
author="Zhu, Wei
and Li, Wenfeng
and Wang, Xiaoling
and Ji, Wendi
and Wu, Yuanbin
and Chen, Jin
and Chen, Liang
and Tang, Buzhou",
editor="Tang, Buzhou
and Chen, Qingcai
and Lin, Hongfei
and Wu, Fei
and Liu, Lei
and Hao, Tianyong
and Wang, Yanshan
and Wang, Haitian
and Lei, Jianbo
and Li, Zuofeng
and Zong, Hui",
title="Extracting Decision Trees from Medical Texts: An Overview of the Text2DT Track in CHIP2022",
booktitle="Health Information Processing. Evaluation Track Papers",
year="2023",
publisher="Springer Nature Singapore",
address="Singapore",
pages="89--102",
isbn="978-981-99-4826-0"
}

@article{Wang2020MiningIH,
  title={Mining Infrequent High-Quality Phrases from Domain-Specific Corpora},
  author={Li Wang and Wei Zhu and Sihang Jiang and Sheng Zhang and Keqiang Wang and Yuan Ni and Guo Tong Xie and Yanghua Xiao},
  journal={Proceedings of the 29th ACM International Conference on Information \& Knowledge Management},
  year={2020},
  url={https://api.semanticscholar.org/CorpusID:224281022}
}

@article{Wang2023MultitaskEL,
  title={Multi-task entity linking with supervision from a taxonomy},
  author={Xuwu Wang and Lihan Chen and Wei Zhu and Yuan Ni and Guo Tong Xie and Deqing Yang and Yanghua Xiao},
  journal={Knowledge and Information Systems},
  year={2023},
  volume={65},
  pages={4335 - 4358},
  url={https://api.semanticscholar.org/CorpusID:258975891}
}

@inproceedings{Zhang2023LearnedAA,
  title={Learned Adapters Are Better Than Manually Designed Adapters},
  author={Yuming Zhang and Peng Wang and Ming Tan and Wei-Guo Zhu},
  booktitle={Annual Meeting of the Association for Computational Linguistics},
  year={2023},
  url={https://api.semanticscholar.org/CorpusID:259858833}
}

@article{Zhu2019TheDS,
  title={The Dr-KGQA System for Automatically Answering Medication Related Questions in Chinese},
  author={Wei Zhu and Yuan Ni and Guo Tong Xie and Xiaofeng Zhou and Cai Chen},
  journal={2019 IEEE International Conference on Healthcare Informatics (ICHI)},
  year={2019},
  pages={1-6},
  url={https://api.semanticscholar.org/CorpusID:208207213}
}

@inproceedings{Zhu2021MVPBERTMP,
  title={MVP-BERT: Multi-Vocab Pre-training for Chinese BERT},
  author={Wei Zhu},
  booktitle={Annual Meeting of the Association for Computational Linguistics},
  year={2021},
  url={https://api.semanticscholar.org/CorpusID:237331564}
}

@inproceedings{Zhang2023LECOIE,
  title={LECO: Improving Early Exiting via Learned Exits and Comparison-based Exiting Mechanism},
  author={Jingfang Zhang and Ming Tan and Pengyu Dai and Wei-Guo Zhu},
  booktitle={Annual Meeting of the Association for Computational Linguistics},
  year={2023},
  url={https://api.semanticscholar.org/CorpusID:259370796}
}

@inproceedings{Zhu2023BADGESU,
  title={BADGE: Speeding Up BERT Inference after Deployment via Block-wise Bypasses and Divergence-based Early Exiting},
  author={Wei Zhu and Peifeng Wang and Yuan Ni and Guo Tong Xie and Xiaoling Wang},
  booktitle={Annual Meeting of the Association for Computational Linguistics},
  year={2023},
  url={https://api.semanticscholar.org/CorpusID:259370582}
}

@article{Zheng2023CandidateSF,
  title={Candidate Soups: Fusing Candidate Results Improves Translation Quality for Non-Autoregressive Translation},
  author={Huanran Zheng and Wei Zhu and Pengfei Wang and Xiaoling Wang},
  journal={ArXiv},
  year={2023},
  volume={abs/2301.11503},
  url={https://api.semanticscholar.org/CorpusID:256358677}
}

@article{Wei2022ChainOT,
  title={Chain of Thought Prompting Elicits Reasoning in Large Language Models},
  author={Jason Wei and Xuezhi Wang and Dale Schuurmans and Maarten Bosma and Ed Huai-hsin Chi and F. Xia and Quoc Le and Denny Zhou},
  journal={ArXiv},
  year={2022},
  volume={abs/2201.11903},
  url={https://api.semanticscholar.org/CorpusID:246411621}
}

@inproceedings{zhu-tan-2023-spt,
    title = "{SPT}: Learning to Selectively Insert Prompts for Better Prompt Tuning",
    author = "Zhu, Wei  and
      Tan, Ming",
    editor = "Bouamor, Houda  and
      Pino, Juan  and
      Bali, Kalika",
    booktitle = "Proceedings of the 2023 Conference on Empirical Methods in Natural Language Processing",
    month = dec,
    year = "2023",
    address = "Singapore",
    publisher = "Association for Computational Linguistics",
    url = "https://aclanthology.org/2023.emnlp-main.727",
    pages = "11862--11878",
}

@ARTICLE{PromptCBLUE,
       author = {{Zhu}, Wei and {Wang}, Xiaoling and {Zheng}, Huanran and {Chen}, Mosha and {Tang}, Buzhou},
        title = "{PromptCBLUE: A Chinese Prompt Tuning Benchmark for the Medical Domain}",
      journal = {arXiv e-prints},
         year = 2023,
        month = oct,
          eid = {arXiv:2310.14151},
        pages = {arXiv:2310.14151},
          doi = {10.48550/arXiv.2310.14151},
archivePrefix = {arXiv},
       eprint = {2310.14151},
 primaryClass = {cs.CL},
       adsurl = {https://ui.adsabs.harvard.edu/abs/2023arXiv231014151Z}
}

@ARTICLE{2023arXiv230318223Z,
       author = {{Zhao}, Wayne Xin and {Zhou}, Kun and {Li}, Junyi and {Tang}, Tianyi and {Wang}, Xiaolei and {Hou}, Yupeng and {Min}, Yingqian and {Zhang}, Beichen and {Zhang}, Junjie and {Dong}, Zican and {Du}, Yifan and {Yang}, Chen and {Chen}, Yushuo and {Chen}, Zhipeng and {Jiang}, Jinhao and {Ren}, Ruiyang and {Li}, Yifan and {Tang}, Xinyu and {Liu}, Zikang and {Liu}, Peiyu and {Nie}, Jian-Yun and {Wen}, Ji-Rong},
        title = "{A Survey of Large Language Models}",
      journal = {arXiv e-prints},
         year = 2023,
        month = mar,
          eid = {arXiv:2303.18223},
        pages = {arXiv:2303.18223},
          doi = {10.48550/arXiv.2303.18223},
archivePrefix = {arXiv},
       eprint = {2303.18223},
 primaryClass = {cs.CL},
       adsurl = {https://ui.adsabs.harvard.edu/abs/2023arXiv230318223Z}
}

@article{Cui2023UltraFeedbackBL,
  title={UltraFeedback: Boosting Language Models with High-quality Feedback},
  author={Ganqu Cui and Lifan Yuan and Ning Ding and Guanming Yao and Wei Zhu and Yuan Ni and Guotong Xie and Zhiyuan Liu and Maosong Sun},
  journal={ArXiv},
  year={2023},
  volume={abs/2310.01377},
  url={https://api.semanticscholar.org/CorpusID:263605623}
}

@inproceedings{zhu_etal_2021_paht,
  title={paht\_nlp @ MEDIQA 2021: Multi-grained Query Focused Multi-Answer Summarization},
  author={Wei Zhu and Yilong He and Ling Chai and Yuanchun Fan and Yuan Ni and Guo Tong Xie and Xiaoling Wang},
  booktitle={Workshop on Biomedical Natural Language Processing},
  year={2021},
  url={https://api.semanticscholar.org/CorpusID:235097590}
}

@article{Zhu2023OverviewOT,
  title={Overview of the PromptCBLUE Shared Task in CHIP2023},
  author={Wei Zhu and Xiaoling Wang and Mosha Chen and Buzhou Tang},
  journal={ArXiv},
  year={2023},
  volume={abs/2312.17522},
  url={https://api.semanticscholar.org/CorpusID:266690968}
}

@inproceedings{Zhang2023FastNERSU,
  title={FastNER: Speeding up Inferences for Named Entity Recognition Tasks},
  author={Yuming Zhang and Xiangxiang Gao and Wei Zhu and Xiaoling Wang},
  booktitle={International Conference on Advanced Data Mining and Applications},
  year={2023},
  url={https://api.semanticscholar.org/CorpusID:265214231}
}

@article{Xu2023ParameterEfficientFM,
  title={Parameter-Efficient Fine-Tuning Methods for Pretrained Language Models: A Critical Review and Assessment},
  author={Lingling Xu and Haoran Xie and Si-Zhao Joe Qin and Xiaohui Tao and Fu Lee Wang},
  journal={ArXiv},
  year={2023},
  volume={abs/2312.12148}
}

@article{Xin2024ParameterEfficientFF,
  title={Parameter-Efficient Fine-Tuning for Pre-Trained Vision Models: A Survey},
  author={Yi Xin and Siqi Luo and Haodi Zhou and Junlong Du and Xiaohong Liu and Yue Fan and Qing Li and Yuntao Du},
  journal={ArXiv},
  year={2024},
  volume={abs/2402.02242}
}

@article{Liu2022FewShotPF,
  title={Few-shot parameter-efficient fine-tuning is better and cheaper than in-context learning},
  author={Liu, Haokun and Tam, Derek and Muqeeth, Mohammed and Mohta, Jay and Huang, Tenghao and Bansal, Mohit and Raffel, Colin A},
  journal={Advances in Neural Information Processing Systems},
  volume={35},
  pages={1950--1965},
  year={2022}
}

@article{achiam2023gpt4,
  title={Gpt-4 technical report},
  author={Achiam, Josh and Adler, Steven and Agarwal, Sandhini and Ahmad, Lama and Akkaya, Ilge and Aleman, Florencia Leoni and Almeida, Diogo and Altenschmidt, Janko and Altman, Sam and Anadkat, Shyamal and others},
  journal={arXiv preprint arXiv:2303.08774},
  year={2023}
}

@article{anil2023palm2,
  title={Palm 2 technical report},
  author={Anil, Rohan and Dai, Andrew M and Firat, Orhan and Johnson, Melvin and Lepikhin, Dmitry and Passos, Alexandre and Shakeri, Siamak and Taropa, Emanuel and Bailey, Paige and Chen, Zhifeng and others},
  journal={arXiv preprint arXiv:2305.10403},
  year={2023}
}

@article{nori2023capabilities,
  title={Capabilities of gpt-4 on medical challenge problems},
  author={Nori, Harsha and King, Nicholas and McKinney, Scott Mayer and Carignan, Dean and Horvitz, Eric},
  journal={arXiv preprint arXiv:2303.13375},
  year={2023}
}

@article{ji2023survey,
  title={Survey of hallucination in natural language generation},
  author={Ji, Ziwei and Lee, Nayeon and Frieske, Rita and Yu, Tiezheng and Su, Dan and Xu, Yan and Ishii, Etsuko and Bang, Ye Jin and Madotto, Andrea and Fung, Pascale},
  journal={ACM Computing Surveys},
  volume={55},
  number={12},
  pages={1--38},
  year={2023},
  publisher={ACM New York, NY}
}

@article{rawte2023survey,
  title={A survey of hallucination in large foundation models},
  author={Rawte, Vipula and Sheth, Amit and Das, Amitava},
  journal={arXiv preprint arXiv:2309.05922},
  year={2023}
}

@article{tian2024opportunities,
  title={Opportunities and challenges for ChatGPT and large language models in biomedicine and health},
  author={Tian, Shubo and Jin, Qiao and Yeganova, Lana and Lai, Po-Ting and Zhu, Qingqing and Chen, Xiuying and Yang, Yifan and Chen, Qingyu and Kim, Won and Comeau, Donald C and others},
  journal={Briefings in Bioinformatics},
  volume={25},
  number={1},
  pages={bbad493},
  year={2024},
  publisher={Oxford University Press}
}

@article{hersh2024search,
  title={Search still matters: information retrieval in the era of generative AI},
  author={Hersh, William},
  journal={Journal of the American Medical Informatics Association},
  pages={ocae014},
  year={2024},
  publisher={Oxford University Press}
}

@article{lewis2020retrieval,
  title={Retrieval-augmented generation for knowledge-intensive nlp tasks},
  author={Lewis, Patrick and Perez, Ethan and Piktus, Aleksandra and Petroni, Fabio and Karpukhin, Vladimir and Goyal, Naman and K{\"u}ttler, Heinrich and Lewis, Mike and Yih, Wen-tau and Rockt{\"a}schel, Tim and others},
  journal={Advances in Neural Information Processing Systems},
  volume={33},
  pages={9459--9474},
  year={2020}
}

@article{gao2023retrieval,
  title={Retrieval-augmented generation for large language models: A survey},
  author={Gao, Yunfan and Xiong, Yun and Gao, Xinyu and Jia, Kangxiang and Pan, Jinliu and Bi, Yuxi and Dai, Yi and Sun, Jiawei and Wang, Haofen},
  journal={arXiv preprint arXiv:2312.10997},
  year={2023}
}

@article{zhao2024retrieval,
  title={Retrieval-Augmented Generation for AI-Generated Content: A Survey},
  author={Zhao, Penghao and Zhang, Hailin and Yu, Qinhan and Wang, Zhengren and Geng, Yunteng and Fu, Fangcheng and Yang, Ling and Zhang, Wentao and Cui, Bin},
  journal={arXiv preprint arXiv:2402.19473},
  year={2024}
}

@article{jin2024agentmd,
  title={AgentMD: Empowering Language Agents for Risk Prediction with Large-Scale Clinical Tool Learning},
  author={Jin, Qiao and Wang, Zhizheng and Yang, Yifan and Zhu, Qingqing and Wright, Donald and Huang, Thomas and Wilbur, W John and He, Zhe and Taylor, Andrew and Chen, Qingyu and others},
  journal={arXiv preprint arXiv:2402.13225},
  year={2024}
}

@article{lala2023paperqa,
  title={Paperqa: Retrieval-augmented generative agent for scientific research},
  author={L{\'a}la, Jakub and O'Donoghue, Odhran and Shtedritski, Aleksandar and Cox, Sam and Rodriques, Samuel G and White, Andrew D},
  journal={arXiv preprint arXiv:2312.07559},
  year={2023}
}

@article{zakka2024almanac,
  title={Almanac—retrieval-augmented language models for clinical medicine},
  author={Zakka, Cyril and Shad, Rohan and Chaurasia, Akash and Dalal, Alex R and Kim, Jennifer L and Moor, Michael and Fong, Robyn and Phillips, Curran and Alexander, Kevin and Ashley, Euan and others},
  journal={NEJM AI},
  volume={1},
  number={2},
  pages={AIoa2300068},
  year={2024},
  publisher={Massachusetts Medical Society}
}

@inproceedings{borgeaud2022improving,
  title={Improving language models by retrieving from trillions of tokens},
  author={Borgeaud, Sebastian and Mensch, Arthur and Hoffmann, Jordan and Cai, Trevor and Rutherford, Eliza and Millican, Katie and Van Den Driessche, George Bm and Lespiau, Jean-Baptiste and Damoc, Bogdan and Clark, Aidan and others},
  booktitle={International conference on machine learning},
  pages={2206--2240},
  year={2022},
  organization={PMLR}
}

@article{zhang2023repocoder,
  title={Repocoder: Repository-level code completion through iterative retrieval and generation},
  author={Zhang, Fengji and Chen, Bei and Zhang, Yue and Keung, Jacky and Liu, Jin and Zan, Daoguang and Mao, Yi and Lou, Jian-Guang and Chen, Weizhu},
  journal={arXiv preprint arXiv:2303.12570},
  year={2023}
}

@article{ram2023context,
  title={In-context retrieval-augmented language models},
  author={Ram, Ori and Levine, Yoav and Dalmedigos, Itay and Muhlgay, Dor and Shashua, Amnon and Leyton-Brown, Kevin and Shoham, Yoav},
  journal={Transactions of the Association for Computational Linguistics},
  volume={11},
  pages={1316--1331},
  year={2023},
  publisher={MIT Press One Broadway, 12th Floor, Cambridge, Massachusetts 02142, USA~…}
}

@article{jiang2023active,
  title={Active retrieval augmented generation},
  author={Jiang, Zhengbao and Xu, Frank F and Gao, Luyu and Sun, Zhiqing and Liu, Qian and Dwivedi-Yu, Jane and Yang, Yiming and Callan, Jamie and Neubig, Graham},
  journal={arXiv preprint arXiv:2305.06983},
  year={2023}
}

@article{zhang2024raft,
  title={Raft: Adapting language model to domain specific rag},
  author={Zhang, Tianjun and Patil, Shishir G and Jain, Naman and Shen, Sheng and Zaharia, Matei and Stoica, Ion and Gonzalez, Joseph E},
  journal={arXiv preprint arXiv:2403.10131},
  year={2024}
}

@article{siriwardhana2023improving,
  title={Improving the domain adaptation of retrieval augmented generation (RAG) models for open domain question answering},
  author={Siriwardhana, Shamane and Weerasekera, Rivindu and Wen, Elliott and Kaluarachchi, Tharindu and Rana, Rajib and Nanayakkara, Suranga},
  journal={Transactions of the Association for Computational Linguistics},
  volume={11},
  pages={1--17},
  year={2023},
  publisher={MIT Press One Broadway, 12th Floor, Cambridge, Massachusetts 02142, USA~…}
}

@article{xue2024badrag,
  title={BadRAG: Identifying Vulnerabilities in Retrieval Augmented Generation of Large Language Models},
  author={Xue, Jiaqi and Zheng, Mengxin and Hu, Yebowen and Liu, Fei and Chen, Xun and Lou, Qian},
  journal={arXiv preprint arXiv:2406.00083},
  year={2024}
}

@article{saab2024capabilities,
  title={Capabilities of gemini models in medicine},
  author={Saab, Khaled and Tu, Tao and Weng, Wei-Hung and Tanno, Ryutaro and Stutz, David and Wulczyn, Ellery and Zhang, Fan and Strother, Tim and Park, Chunjong and Vedadi, Elahe and others},
  journal={arXiv preprint arXiv:2404.18416},
  year={2024}
}

@article{chen2023large,
  title={Large language models in biomedical natural language processing: benchmarks, baselines, and recommendations},
  author={Chen, Qingyu and Du, Jingcheng and Hu, Yan and Keloth, Vipina Kuttichi and Peng, Xueqing and Raja, Kalpana and Zhang, Rui and Lu, Zhiyong and Xu, Hua},
  journal={arXiv preprint arXiv:2305.16326},
  year={2023}
}

@inproceedings{frisoni2022bioreader,
  title={Bioreader: a retrieval-enhanced text-to-text transformer for biomedical literature},
  author={Frisoni, Giacomo and Mizutani, Miki and Moro, Gianluca and Valgimigli, Lorenzo},
  booktitle={Proceedings of the 2022 conference on empirical methods in natural language processing},
  pages={5770--5793},
  year={2022}
}

@article{naik2021literature,
  title={Literature-augmented clinical outcome prediction},
  author={Naik, Aakanksha and Parasa, Sravanthi and Feldman, Sergey and Wang, Lucy Lu and Hope, Tom},
  journal={arXiv preprint arXiv:2111.08374},
  year={2021}
}

@article{xiong2024benchmarking,
  title={Benchmarking retrieval-augmented generation for medicine},
  author={Xiong, Guangzhi and Jin, Qiao and Lu, Zhiyong and Zhang, Aidong},
  journal={arXiv preprint arXiv:2402.13178},
  year={2024}
}

@article{jeong2024improving,
  title={Improving Medical Reasoning through Retrieval and Self-Reflection with Retrieval-Augmented Large Language Models},
  author={Jeong, Minbyul and Sohn, Jiwoong and Sung, Mujeen and Kang, Jaewoo},
  journal={arXiv preprint arXiv:2401.15269},
  year={2024}
}

@article{wang2023augmenting,
  title={Augmenting black-box llms with medical textbooks for clinical question answering},
  author={Wang, Yubo and Ma, Xueguang and Chen, Wenhu},
  journal={arXiv preprint arXiv:2309.02233},
  year={2023}
}

@inproceedings{zweigenbaum2003question,
  title={Question answering in biomedicine},
  author={Zweigenbaum, Pierre},
  booktitle={Proceedings Workshop on Natural Language Processing for Question Answering, EACL},
  volume={2005},
  pages={1--4},
  year={2003},
  organization={Citeseer}
}

@article{athenikos2010biomedical,
  title={Biomedical question answering: A survey},
  author={Athenikos, Sofia J and Han, Hyoil},
  journal={Computer methods and programs in biomedicine},
  volume={99},
  number={1},
  pages={1--24},
  year={2010},
  publisher={Elsevier}
}

@article{wang2023cmb,
  title={Cmb: A comprehensive medical benchmark in chinese},
  author={Wang, Xidong and Chen, Guiming Hardy and Song, Dingjie and Zhang, Zhiyi and Chen, Zhihong and Xiao, Qingying and Jiang, Feng and Li, Jianquan and Wan, Xiang and Wang, Benyou and others},
  journal={arXiv preprint arXiv:2308.08833},
  year={2023}
}

@article{suzgun2022challenging,
  title={Challenging big-bench tasks and whether chain-of-thought can solve them},
  author={Suzgun, Mirac and Scales, Nathan and Sch{\"a}rli, Nathanael and Gehrmann, Sebastian and Tay, Yi and Chung, Hyung Won and Chowdhery, Aakanksha and Le, Quoc V and Chi, Ed H and Zhou, Denny and others},
  journal={arXiv preprint arXiv:2210.09261},
  year={2022}
}

@article{yue2024tcmbench,
  title={TCMBench: A Comprehensive Benchmark for Evaluating Large Language Models in Traditional Chinese Medicine},
  author={Yue, Wenjing and Wang, Xiaoling and Zhu, Wei and Guan, Ming and Zheng, Huanran and Wang, Pengfei and Sun, Changzhi and Ma, Xin},
  journal={arXiv preprint arXiv:2406.01126},
  year={2024}
}

@inproceedings{pal2022medmcqa,
  title={Medmcqa: A large-scale multi-subject multi-choice dataset for medical domain question answering},
  author={Pal, Ankit and Umapathi, Logesh Kumar and Sankarasubbu, Malaikannan},
  booktitle={Conference on health, inference, and learning},
  pages={248--260},
  year={2022},
  organization={PMLR}
}

@article{jin2019pubmedqa,
  title={Pubmedqa: A dataset for biomedical research question answering},
  author={Jin, Qiao and Dhingra, Bhuwan and Liu, Zhengping and Cohen, William W and Lu, Xinghua},
  journal={arXiv preprint arXiv:1909.06146},
  year={2019}
}

@article{krithara2023bioasq,
  title={BioASQ-QA: A manually curated corpus for Biomedical Question Answering},
  author={Krithara, Anastasia and Nentidis, Anastasios and Bougiatiotis, Konstantinos and Paliouras, Georgios},
  journal={Scientific Data},
  volume={10},
  number={1},
  pages={170},
  year={2023},
  publisher={Nature Publishing Group UK London}
}

@article{jin2020disease,
  title={What disease does this patient have},
  author={Jin, Di and Pan, Eileen and Oufattole, Nassim and Weng, Wei-Hung and Fang, Hanyi and Szolovits, Peter},
  journal={A Large-scale Open Domain Question Answering Dataset from Medical Exams. arXiv [cs. CL]},
  year={2020}
}

@article{xue2003studying,
  title={Studying traditional Chinese medicine},
  author={Xue, Tianhan and Roy, Rustum},
  journal={Science},
  volume={300},
  number={5620},
  pages={740--741},
  year={2003},
  publisher={American Association for the Advancement of Science}
}

@inproceedings{abacha2017overview,
  title={Overview of the medical question answering task at TREC 2017 LiveQA.},
  author={Abacha, Asma Ben and Agichtein, Eugene and Pinter, Yuval and Demner-Fushman, Dina},
  booktitle={TREC},
  pages={1--12},
  year={2017}
}

@inproceedings{abacha2019bridging,
  title={Bridging the Gap Between Consumers' Medication Questions and Trusted Answers.},
  author={Abacha, Asma Ben and Mrabet, Yassine and Sharp, Mark and Goodwin, Travis R and Shooshan, Sonya E and Demner-Fushman, Dina},
  booktitle={MedInfo},
  pages={25--29},
  year={2019}
}

@article{herrero2013ddi,
  title={The DDI corpus: An annotated corpus with pharmacological substances and drug--drug interactions},
  author={Herrero-Zazo, Mar{\'\i}a and Segura-Bedmar, Isabel and Mart{\'\i}nez, Paloma and Declerck, Thierry},
  journal={Journal of biomedical informatics},
  volume={46},
  number={5},
  pages={914--920},
  year={2013},
  publisher={Elsevier}
}

@article{taboureau2010chemprot,
  title={ChemProt: a disease chemical biology database},
  author={Taboureau, Olivier and Nielsen, Sonny Kim and Audouze, Karine and Weinhold, Nils and Edsg{\"a}rd, Daniel and Roque, Francisco S and Kouskoumvekaki, Irene and Bora, Alina and Curpan, Ramona and Jensen, Thomas Sk{\o}t and others},
  journal={Nucleic acids research},
  volume={39},
  number={suppl\_1},
  pages={D367--D372},
  year={2010},
  publisher={Oxford University Press}
}

@inproceedings{guan2020cmeie,
  title={CMeIE: construction and evaluation of Chinese medical information extraction dataset},
  author={Guan, Tongfeng and Zan, Hongying and Zhou, Xiabing and Xu, Hongfei and Zhang, Kunli},
  booktitle={Natural Language Processing and Chinese Computing: 9th CCF International Conference, NLPCC 2020, Zhengzhou, China, October 14--18, 2020, Proceedings, Part I 9},
  pages={270--282},
  year={2020},
  organization={Springer}
}

@article{kumar2020link,
  title={Link prediction techniques, applications, and performance: A survey},
  author={Kumar, Ajay and Singh, Shashank Sheshar and Singh, Kuldeep and Biswas, Bhaskar},
  journal={Physica A: Statistical Mechanics and its Applications},
  volume={553},
  pages={124289},
  year={2020},
  publisher={Elsevier}
}

@inproceedings{aruna2022survey,
  title={A Survey of Recent Techniques in Computational Drug Repurposing},
  author={Aruna, AS and Remesh Babu, KR and Deepthi, K},
  booktitle={International Conference on Intelligent Systems Design and Applications},
  pages={565--575},
  year={2022},
  organization={Springer}
}

@article{xiao2024repurposing,
  title={Repurposing non-pharmacological interventions for Alzheimer's disease through link prediction on biomedical literature},
  author={Xiao, Yongkang and Hou, Yu and Zhou, Huixue and Diallo, Gayo and Fiszman, Marcelo and Wolfson, Julian and Zhou, Li and Kilicoglu, Halil and Chen, You and Su, Chang and others},
  journal={Scientific Reports},
  volume={14},
  number={1},
  pages={8693},
  year={2024},
  publisher={Nature Publishing Group UK London}
}

@article{ioannidis2020drkg,
  title={Drkg-drug repurposing knowledge graph for covid-19},
  author={Ioannidis, Vassilis N and Song, Xiang and Manchanda, Saurav and Li, Mufei and Pan, Xiaoqin and Zheng, Da and Ning, Xia and Zeng, Xiangxiang and Karypis, George},
  journal={arXiv preprint arXiv:2010.09600},
  year={2020}
}

@article{chiang2024chatbot,
  title={Chatbot arena: An open platform for evaluating llms by human preference},
  author={Chiang, Wei-Lin and Zheng, Lianmin and Sheng, Ying and Angelopoulos, Anastasios Nikolas and Li, Tianle and Li, Dacheng and Zhang, Hao and Zhu, Banghua and Jordan, Michael and Gonzalez, Joseph E and others},
  journal={arXiv preprint arXiv:2403.04132},
  year={2024}
}

@article{robertson2009probabilistic,
  title={The probabilistic relevance framework: BM25 and beyond},
  author={Robertson, Stephen and Zaragoza, Hugo and others},
  journal={Foundations and Trends{\textregistered} in Information Retrieval},
  volume={3},
  number={4},
  pages={333--389},
  year={2009},
  publisher={Now Publishers, Inc.}
}

@article{jin2023medcpt,
  title={MedCPT: Contrastive Pre-trained Transformers with large-scale PubMed search logs for zero-shot biomedical information retrieval},
  author={Jin, Qiao and Kim, Won and Chen, Qingyu and Comeau, Donald C and Yeganova, Lana and Wilbur, W John and Lu, Zhiyong},
  journal={Bioinformatics},
  volume={39},
  number={11},
  pages={btad651},
  year={2023},
  publisher={Oxford University Press}
}

@article{xiao2023c,
  title={C-pack: Packaged resources to advance general chinese embedding},
  author={Xiao, Shitao and Liu, Zheng and Zhang, Peitian and Muennighof, Niklas},
  journal={arXiv preprint arXiv:2309.07597},
  year={2023}
}

@article{wang2023improving,
  title={Improving text embeddings with large language models},
  author={Wang, Liang and Yang, Nan and Huang, Xiaolong and Yang, Linjun and Majumder, Rangan and Wei, Furu},
  journal={arXiv preprint arXiv:2401.00368},
  year={2023}
}

@inproceedings{cormack2009reciprocal,
  title={Reciprocal rank fusion outperforms condorcet and individual rank learning methods},
  author={Cormack, Gordon V and Clarke, Charles LA and Buettcher, Stefan},
  booktitle={Proceedings of the 32nd international ACM SIGIR conference on Research and development in information retrieval},
  pages={758--759},
  year={2009}
}

@article{bai2023qwen,
  title={Qwen technical report},
  author={Bai, Jinze and Bai, Shuai and Chu, Yunfei and Cui, Zeyu and Dang, Kai and Deng, Xiaodong and Fan, Yang and Ge, Wenbin and Han, Yu and Huang, Fei and others},
  journal={arXiv preprint arXiv:2309.16609},
  year={2023}
}

@article{chen2023meditron,
  title={Meditron-70b: Scaling medical pretraining for large language models},
  author={Chen, Zeming and Cano, Alejandro Hern{\'a}ndez and Romanou, Angelika and Bonnet, Antoine and Matoba, Kyle and Salvi, Francesco and Pagliardini, Matteo and Fan, Simin and K{\"o}pf, Andreas and Mohtashami, Amirkeivan and others},
  journal={arXiv preprint arXiv:2311.16079},
  year={2023}
}

@article{wu2023pmc,
  title={Pmc-llama: Further finetuning llama on medical papers},
  author={Wu, Chaoyi and Zhang, Xiaoman and Zhang, Ya and Wang, Yanfeng and Xie, Weidi},
  journal={arXiv preprint arXiv:2304.14454},
  year={2023}
}

@article{bao2023disc,
  title={Disc-medllm: Bridging general large language models and real-world medical consultation},
  author={Bao, Zhijie and Chen, Wei and Xiao, Shengze and Ren, Kuang and Wu, Jiaao and Zhong, Cheng and Peng, Jiajie and Huang, Xuanjing and Wei, Zhongyu},
  journal={arXiv preprint arXiv:2308.14346},
  year={2023}
}

@article{madaan2024self,
  title={Self-refine: Iterative refinement with self-feedback},
  author={Madaan, Aman and Tandon, Niket and Gupta, Prakhar and Hallinan, Skyler and Gao, Luyu and Wiegreffe, Sarah and Alon, Uri and Dziri, Nouha and Prabhumoye, Shrimai and Yang, Yiming and others},
  journal={Advances in Neural Information Processing Systems},
  volume={36},
  year={2024}
}

@article{glickman2010uscf,
  title={The USCF rating system},
  author={Glickman, Mark E and Doan, Thomas},
  journal={URL http://www. glicko. net/ratings/rating. system. pdf},
  year={2010}
}

@article{kaplan2020scaling,
  title={Scaling laws for neural language models},
  author={Kaplan, Jared and McCandlish, Sam and Henighan, Tom and Brown, Tom B and Chess, Benjamin and Child, Rewon and Gray, Scott and Radford, Alec and Wu, Jeffrey and Amodei, Dario},
  journal={arXiv preprint arXiv:2001.08361},
  year={2020}
}

@article{reid2024gemini,
  title={Gemini 1.5: Unlocking multimodal understanding across millions of tokens of context},
  author={Reid, Machel and Savinov, Nikolay and Teplyashin, Denis and Lepikhin, Dmitry and Lillicrap, Timothy and Alayrac, Jean-baptiste and Soricut, Radu and Lazaridou, Angeliki and Firat, Orhan and Schrittwieser, Julian and others},
  journal={arXiv preprint arXiv:2403.05530},
  year={2024}
}

@article{thakur2021beir,
  title={Beir: A heterogenous benchmark for zero-shot evaluation of information retrieval models},
  author={Thakur, Nandan and Reimers, Nils and R{\"u}ckl{\'e}, Andreas and Srivastava, Abhishek and Gurevych, Iryna},
  journal={arXiv preprint arXiv:2104.08663},
  year={2021}
}

@article{chen2017reading,
  title={Reading wikipedia to answer open-domain questions},
  author={Chen, Danqi and Fisch, Adam and Weston, Jason and Bordes, Antoine},
  journal={arXiv preprint arXiv:1704.00051},
  year={2017}
}

@article{zheng2024judging,
  title={Judging llm-as-a-judge with mt-bench and chatbot arena},
  author={Zheng, Lianmin and Chiang, Wei-Lin and Sheng, Ying and Zhuang, Siyuan and Wu, Zhanghao and Zhuang, Yonghao and Lin, Zi and Li, Zhuohan and Li, Dacheng and Xing, Eric and others},
  journal={Advances in Neural Information Processing Systems},
  volume={36},
  year={2024}
}

@article{holtzman2019curious,
  title={The curious case of neural text degeneration},
  author={Holtzman, Ari and Buys, Jan and Du, Li and Forbes, Maxwell and Choi, Yejin},
  journal={arXiv preprint arXiv:1904.09751},
  year={2019}
}

@article{wang2024ts,
  title={TS-TCD: Triplet-Level Cross-Modal Distillation for Time-Series Forecasting Using Large Language Models},
  author={Wang, Pengfei and Zheng, Huanran and Dai, Silong and Yue, Wenjing and Zhu, Wei and Wang, Xiaoling},
  journal={arXiv preprint arXiv:2409.14978},
  year={2024}
}

@article{zhu2024iapt,
  title={IAPT: Instruction-Aware Prompt Tuning for Large Language Models},
  author={Zhu, Wei and Tian, Aaron Xuxiang and Yin, Congrui and Ni, Yuan and Wang, Xiaoling and Xie, Guotong},
  journal={arXiv preprint arXiv:2405.18203},
  year={2024}
}

@article{xie2024pedro,
  title={PEDRO: Parameter-Efficient Fine-tuning with Prompt DEpenDent Representation MOdification},
  author={Xie, Tianfang and Li, Tianjing and Zhu, Wei and Han, Wei and Zhao, Yi},
  journal={arXiv preprint arXiv:2409.17834},
  year={2024}
}

@inproceedings{zheng2024nat4at,
  title={NAT4AT: Using Non-Autoregressive Translation Makes Autoregressive Translation Faster and Better},
  author={Zheng, Huanran and Zhu, Wei and Wang, Xiaoling},
  booktitle={Proceedings of the ACM on Web Conference 2024},
  pages={4181--4192},
  year={2024}
}

@inproceedings{zhu2023acf,
  title={ACF: aligned contrastive finetuning for language and vision tasks},
  author={Zhu, Wei and Wang, Peng and Wang, Xiaoling and Ni, Yuan and Xie, Guotong},
  booktitle={ICASSP 2023-2023 IEEE International Conference on Acoustics, Speech and Signal Processing (ICASSP)},
  pages={1--5},
  year={2023},
  organization={IEEE}
}

@inproceedings{zhu2019panlp,
  title={Panlp at mediqa 2019: Pre-trained language models, transfer learning and knowledge distillation},
  author={Zhu, Wei and Zhou, Xiaofeng and Wang, Keqiang and Luo, Xun and Li, Xiepeng and Ni, Yuan and Xie, Guotong},
  booktitle={Proceedings of the 18th BioNLP Workshop and Shared Task},
  pages={380--388},
  year={2019}
}

@inproceedings{zhu2019dr,
  title={The DR-KGQA system for automatically answering medication related questions in Chinese},
  author={Zhu, Wei and Ni, Yuan and Xie, Guotong and Zhou, Xiaofeng and Chen, Cai},
  booktitle={2019 IEEE International Conference on Healthcare Informatics (ICHI)},
  pages={1--6},
  year={2019},
  organization={IEEE}
}

@incollection{zhou2019analysis,
  title={Analysis of the health information needs of diabetics in China},
  author={Zhou, Xiaofeng and Ni, Yuan and Xie, Guotong and Zhu, Wei and Chen, Cai and Wang, Tianhao and Pan, Zhigang},
  booktitle={MEDINFO 2019: Health and Wellbeing e-Networks for All},
  pages={487--491},
  year={2019},
  publisher={IOS Press}
}

@inproceedings{zhang2025time,
  title={Time-llama: Adapting large language models for time series modeling via dynamic low-rank adaptation},
  author={Zhang, Juyuan and Gao, Jiechao and Ouyang, Wenwen and Zhu, Wei and Leong, Hui Yi},
  booktitle={Proceedings of the 63rd Annual Meeting of the Association for Computational Linguistics (Volume 4: Student Research Workshop)},
  pages={1145--1157},
  year={2025}
}

@article{wang2025ts,
  title={TS-HTFA: Advancing Time-Series Forecasting via Hierarchical Text-Free Alignment with Large Language Models},
  author={Wang, Pengfei and Zheng, Huanran and Xu, Qi’ao and Dai, Silong and Wang, Yiqiao and Yue, Wenjing and Zhu, Wei and Qian, Tianwen and Zhao, Liang},
  journal={Symmetry},
  volume={17},
  number={3},
  pages={401},
  year={2025},
  publisher={MDPI}
}

@article{liu2025parameter,
  title={PARA: Parameter-Efficient Fine-tuning with Prompt Aware Representation Adjustment},
  author={Liu, Zequan and Zhao, Yi and Tan, Ming and Zhu, Wei and Tian, Aaron Xuxiang},
  journal={arXiv preprint arXiv:2502.01033},
  year={2025}
}

@article{yi2024drum,
  title={DRUM: Learning Demonstration Retriever for Large MUlti-modal Models},
  author={Yi-Ge, Ellen and Gao, Jiechao and Han, Wei and Zhu, Wei},
  journal={arXiv preprint arXiv:2412.07619},
  year={2024}
}

@inproceedings{tian2024fanlora,
  title={Fanlora: Fantastic loras and where to find them in large language model fine-tuning},
  author={Tian, Aaron and Zhao, Yi and Yin, Congrui and Zhu, Wei and Tian, Xing and Ge, Yi},
  booktitle={Proceedings of the 2024 Conference on Empirical Methods in Natural Language Processing: Industry Track},
  pages={515--528},
  year={2024}
}

@inproceedings{li2025ft,
  title={FT-MDT: Extracting Decision Trees from Medical Texts via a Novel Low-rank Adaptation Method},
  author={Li, Yuheng and Gao, Jiechao and Han, Wei and Ouyang, Wenwen and Zhu, Wei and Leong, Hui Yi},
  booktitle={Proceedings of the 2025 Conference on Empirical Methods in Natural Language Processing: Industry Track},
  pages={65--76},
  year={2025}
}

@inproceedings{leong2025amas,
  title={Amas: Adaptively determining communication topology for llm-based multi-agent system},
  author={Leong, Hui Yi and Li, Yuheng and Wu, Yuqing and Ouyang, Wenwen and Zhu, Wei and Gao, Jiechao and Han, Wei},
  booktitle={Proceedings of the 2025 Conference on Empirical Methods in Natural Language Processing: Industry Track},
  pages={2061--2070},
  year={2025}
}

@article{yin2024machine,
  title={A machine learning model for predicting acute exacerbation of in-home chronic obstructive pulmonary disease patients},
  author={Yin, Huiming and Wang, Kun and Yang, Ruyu and Tan, Yanfang and Li, Qiang and Zhu, Wei and Sung, Suzi},
  journal={Computer Methods and Programs in Biomedicine},
  volume={246},
  pages={108005},
  year={2024},
  publisher={Elsevier}
}

@inproceedings{zhu2021leebert,
  title={LeeBERT: Learned early exit for BERT with cross-level optimization},
  author={Zhu, Wei},
  booktitle={Proceedings of the 59th Annual Meeting of the Association for Computational Linguistics and the 11th International Joint Conference on Natural Language Processing (Volume 1: Long Papers)},
  pages={2968--2980},
  year={2021}
}

@inproceedings{zheng2024sca,
  title={Sca: Selective compression attention for efficiently extending the context window of large language models},
  author={Zheng, Huanran and Zhu, Wei and Wang, Xiaoling},
  booktitle={Findings of the Association for Computational Linguistics: EMNLP 2024},
  pages={6166--6178},
  year={2024}
}

@article{zhang2024milora,
  title={MiLoRA: Efficient mixture of low-rank adaptation for large language models fine-tuning},
  author={Zhang, Jingfan and Zhao, Yi and Chen, Dan and Tian, Xing and Zheng, Huanran and Zhu, Wei},
  journal={arXiv preprint arXiv:2410.18035},
  year={2024}
}

\appendix

\section{Appendix: Related works}
\label{sec:app_related_works}

\subsection{Question answering in bio-medicine}

In the LLM era, almost all the bio-medical information needs are expressed as natural language questions \cite{zweigenbaum2003question,athenikos2010biomedical,PromptCBLUE,Text2dt}, such as user queries about healthcare, or searches for specific knowledge from medical practitioners, or the need to make the knowledge structural from an unstructured document for bio-medical knowledge graph construction. LLMs, both open-domain and domain-specific, have demonstrated outstanding potential for medical QA tasks \cite{achiam2023gpt4,anil2023palm2,singhal2023large,singhal2023towards,nori2023capabilities,chen2023large,saab2024capabilities}, with chain-of-thought prompting or in-context learning. However, due to the knowledge-intensive nature, intuitively, RAG could help the LLMs achieve better performance and be more reliable by grounding their responses to the retrieved referential documents. In this work, our proposed MRAG-Bench consists of four cohorts of bio-medical tasks, reflecting different information-seeking needs in this domain. Through experiments, we can see that some of the MRAG tasks are challenging for LLMs, even with the help of RAG.

\section{MRAG-Bench datasets}
\label{sec:app_mrag_bench_datasets}

In this section, we provide detailed introductions, statistics, filtering procedures, and licensing information for all the tasks in the MRAG-Bench. Our MRAG-Bench consists of 13 tasks, 9 of which were previously open-sourced by their original authors. We construct two link prediction datasets from open-sourced knowledge graphs and curate two novel Chinese datasets.

\subsection{Previously open-sourced datasets}

\textbf{MMLU-Med.} Massive Multitask Language Understanding (MMLU) \cite{hendrycks2020measuring} is a benchmark for the evaluation of the multitask learning capability of language models. The dataset is released under the MIT License and in \url{https://github.com/hendrycks/test}. The benchmark contains a variety of 57 different tasks. To
measure the performance of medical RAG systems, we select six tasks related to biomedicine following \cite{singhal2023large}, including anatomy, clinical knowledge, professional medicine, human genetics, college medicine, and
college biology. The subset is collectively denoted as MMLU-Med. Only the test set of each task is used in our benchmark, which contains 1089 questions in total.

\textbf{MedQA-US.} MedQA \cite{jin2020disease} is a multi-choice question answering (MCQA) dataset collected from professional medical board exams. The dataset is released under the CC BY 4.0 (Creative Commons Attribution 4.0 International) license and released in \url{https://github.com/jind11/MedQA}. Specifically, we focus on the English part, which includes real-world questions from the US Medical Licensing Examination (MedQA-US). Thus, the subset of 1273 four-option test questions are included in our MRAG-Bench.

\textbf{MedMCQA.} MedMCQA \cite{pal2022medmcqa} contains 194k multi-choice questions collected from Indian medical entrance exams. The dataset is released under the MIT License and in \url{https://medmcqa.github.io/}. The questions cover a range of 2.4k healthcare topics and 21 medical subjects. Since the ground truth of its test set is not provided, the dev set of the original MedMCQA is chosen for MRAG-Bench, including 4183 medical questions. To ensure balance in the task composition, we randomly selected 1,500 test samples from MedMCQA's dev set.

\textbf{PubMedQA.} PubMedQA \cite{jin2019pubmedqa} is a biomedical research QA dataset. The dataset is released under the CC BY 4.0 (Creative Commons Attribution 4.0 International) license and released in \url{https://pubmedqa.github.io/}. It has 1k manually annotated questions constructed from PubMed abstracts. To test the capability of RAG systems to find related documents and answer the question accordingly, we discard the relevant context for each question originally included in the dataset. A test set of 500 PubMedQA samples is adopted for MRAG-Bench following \cite{lala2023paperqa}. The possible answer to a PubMedQA question can be yes/no/maybe, reflecting the authenticity of the question statement based on biomedical literature.

\textbf{BioASQ.} BioASQ \cite{krithara2023bioasq} is an annual competition for biomedical QA, which includes both the information retrieval track (Task A) and the machine reading comprehension track (Task B). The dataset is released under the Creative Commons Attribution-NonCommercial-ShareAlike 4.0 International License (CC BY-NC-SA 4.0) and released in \url{http://bioasq.org/}. To leverage the resources of BioASQ for our medical RAG benchmark, we select the Yes/No questions in the ground truth test set of Task B from the most recent five years (2019-2023), including 618 questions. In the original task, questions are constructed based on biomedical literature, and the ground truth document snippets are provided as a basis for machine reading comprehension. We discard the provided document snippets and only keep the questions and answer choices for the MRAG-Bench.

\textbf{ChemProt.} ChemProt \cite{taboureau2010chemprot} is a specialized task in the field of bio-medical information extraction (IE), focusing on extracting interactions among diseases, chemical compounds, and genes from medical articles. This task is essential for understanding biochemical processes and advancing drug discovery and development. The ChemProt corpus is distributed under a Creative Commons Attribution-NonCommercial-ShareAlike 3.0 Unported License, and it is publicly available at \url{https://biocreative.bioinformatics.udel.edu/news/corpora/chemprot-corpus-biocreative-vi/}. For MRAG-Bench, we include its test set, which consists of 800 samples.

\textbf{DDI.} The DDI Extraction (DDI) 2013 task \cite{herrero2013ddi} is a significant benchmark in the field of bio-medical natural language processing (NLP), focusing on the extraction of drug-drug interactions (DDIs) from textual data. The task is structured to challenge participants to develop and refine algorithms capable of accurately parsing complex biomedical texts to detect and categorize these interactions. This task is crucial for improving our understanding of how different drugs interact, vital for drug safety, patient care, and the development of new pharmaceutical treatments. This task relies on the DDI corpus, which includes MedLine abstracts and documents from the DrugBank database. These documents have been manually annotated with pharmacological substances and drug-drug interactions. The DDI corpus is licensed under the Creative Commons Attribution-NonCommercial 4.0 International (CC BY-NC 4.0) license and is available at \url{https://github.com/isegura/DDICorpus}.

\textbf{CMeIE.} The Chinese Medical Information Extraction (CMeIE) dataset \cite{guan2020cmeie}, part of the CHIP-2020 shared tasks\footnote{\url{http://cips-chip.org.cn/2020/eval2}}, is a crucial resource for advancing Chinese natural language processing (NLP) in the medical domain. This dataset is designed to facilitate medical information extraction by identifying entities and their relationships within clinical text, following predefined schema constraints. This dataset is released under the Creative Commons Attribution 4.0 International (CC BY 4.0) License and is available at \url{http://biendata.com/competition/chip_2020_2}. Since the original authors keep its test set private, we randomly sample a 1,500-sample subset from its dev set for MRAG-Bench.

\textbf{MultiMedQA.} The MultiMedQA dataset \cite{singhal2023towards} is a comprehensive collection of medical question sets designed to support the development and evaluation of long-form question-answering (LFQA) systems in the healthcare domain. This dataset is constructed by combining 1066 questions curated from three distinct sources: HealthSearchQA \cite{singhal2023large}, LiveQA \cite{abacha2017overview}, and MedicationQA \cite{abacha2019bridging}. The diverse nature of these sources ensures a wide coverage of medical topics, ranging from general health inquiries to specific medication-related questions. The MultiMedQA dataset is released under the Creative Commons Attribution-NonCommercial 4.0 International (CC BY-NC 4.0) license and is available at \url{https://huggingface.co/datasets/katielink/healthsearchqa/tree/main}.

\subsection{Construction of the two link prediction tasks}

\textbf{Source.} Our link prediction (LP) tasks are constructed based on two open-sourced knowledge graphs (KGs), ADInt \cite{xiao2024repurposing} and DRKG \cite{ioannidis2020drkg}. 
\begin{itemize}
\item \textbf{ADInt.} ADInt is a comprehensive knowledge graph designed to facilitate the identification of novel pharmaceutical interventions (PIs) for Alzheimer's Disease (AD). ADInt aims to accelerate research and discovery in AD therapeutics by integrating and organizing diverse biomedical data. ADInt is distributed under the Creative Commons Attribution 4.0 International (CC BY 4.0) license and is available at \url{https://github.com/zhang-informatics/ADInt}. ADInt harnesses the power of a knowledge graph to map complex relationships between various entities related to Alzheimer's Disease. These entities include genes, proteins, biological pathways, drugs, clinical trials, and research publications. Researchers can identify novel drug candidates and repurposing opportunities by exploring connections between known compounds and AD-related targets.

\item \textbf{DRKG.} The Drug Repurposing Knowledge Graph (DRKG) is a comprehensive drug discovery and repurposing research resource. DRKG integrates information from various biological databases to create a unified knowledge graph that includes genes, diseases, drugs, biological processes, and molecular functions. By representing these relationships in a structured format, DRKG enables researchers to identify potential drug repurposing opportunities, uncover novel therapeutic targets, and better understand the complex interactions within biological systems. DRKG is released under the MIT License and is available at \url{https://github.com/gnn4dr/DRKG}.

\end{itemize}

\textbf{Dataset Collection.} In both ADInt and DRKG, a piece of knowledge is expressed as a triplet (subject, predicate, object), in which the subject and object are entities from the knowledge graphs, and the predicate represents the type of relation between the two entities. The relation type is pre-defined as the schema of the KG. We randomly select 1,500 triplets from each of the two KGs to evaluate large language models in MRAG-Bench.

\textbf{Formatting.} For constructing a link prediction task for large language models, a triplet is then formatted into a standardized multi-choice format. The question uses the following template to include the subject and object's names, 
\begin{lstlisting} 
How are the following entity pairs connected? 
Entity 1: <ent1>
Entity 2: <ent2>
\end{lstlisting}
Moreover, the correct answer is the predicate name of the triplet. The distractors (incorrect options) are the other relation types in the knowledge graphs. 

\begin{figure*}
\centering
\includegraphics[width=0.84\textwidth]{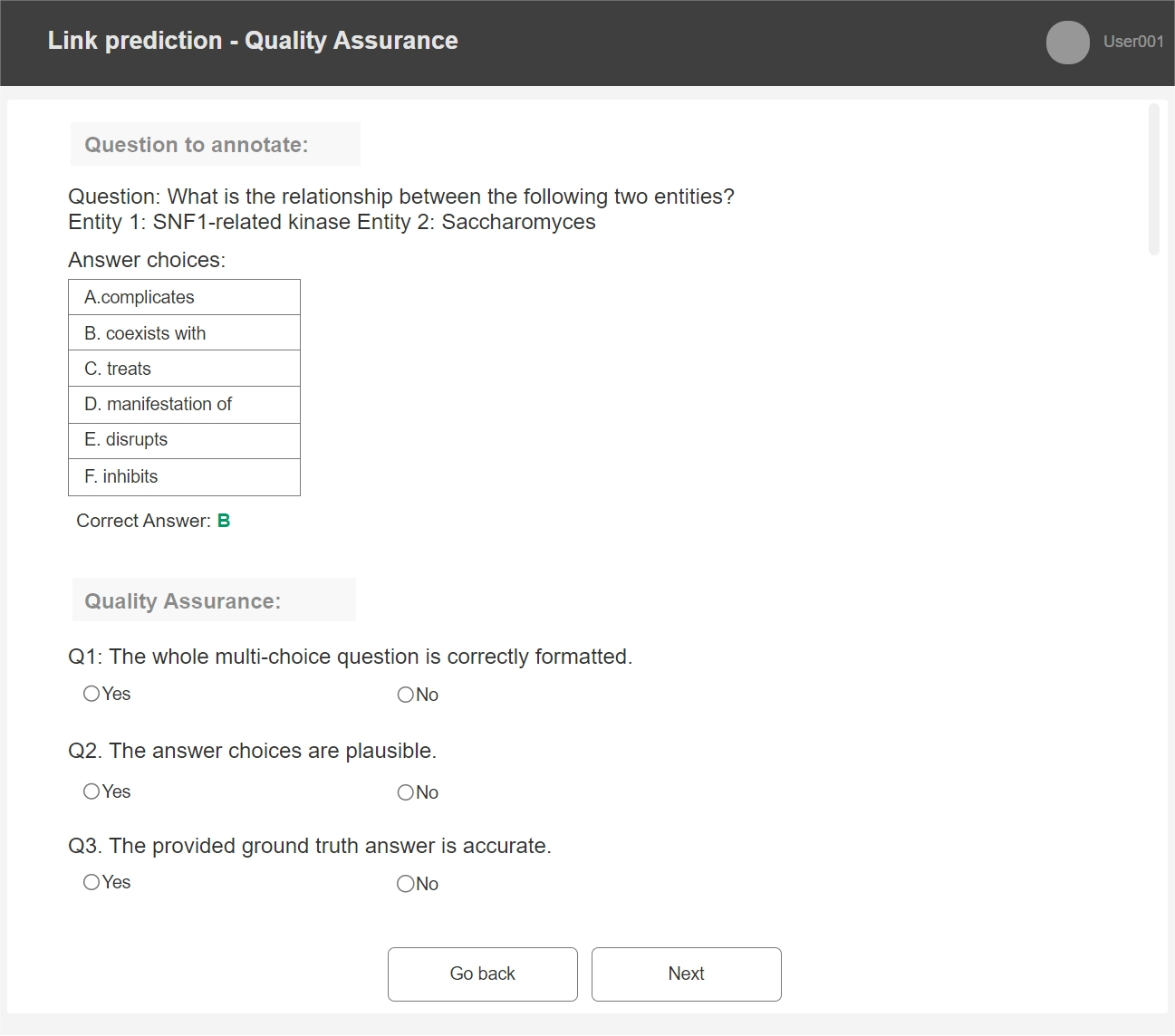} 
\caption{The screenshot of the annotation web-page for quality assurance of the link prediction tasks. }
\label{fig:screen_shot_link_prediction_annotation}
\end{figure*}

\textbf{Quality Assurance.} The dataset undergoes a final review for quality assurance. A pool of 15 medical experts from the united states with medical doctoral degrees and different fields is divided into five groups, each containing three experts. A group will be given a link prediction question in the multi-choice format. These experts participate in this project as volunteers and are paid 10 US dollars per hour. They will check whether (a) the whole question is correctly formatted, (b) the answer choices are plausible, and (c) the correct answers are accurate. The screenshot of the annotation webpage is presented in Figure \ref{fig:screen_shot_link_prediction_annotation}. If not all of the experts in the group agree that the question is valid in all the above three aspects, this question will be filtered out. After the review process, no question is discarded. This result also reflects the high quality of the source KGs.

\textbf{Dataset Compilation.} After the quality assurance process, the questions are compiled into a structured dataset. Each entry in the dataset includes the question text, the answer choices, and the correct answer label.

\subsection{Curation of MRAG-TCM}

For Chinese MCQA, since the exam questions of Western medicine are well covered by the MMLU-Med, MedQA-US, and MedMCQA tasks, we construct an MCQA dataset containing 1200 test samples for Traditional Chinese Medicine (TCM) \cite{xue2003studying}, and refer to this dataset as MRAG-TCM. We now describe how we construct the MRAG-TCM dataset.

\textbf{Source.} The MRAG-TCM dataset is meticulously designed to evaluate the performance of large-scale language models in the context of the Traditional Chinese Medicine Practitioners Qualification Examination (TCMPQE)\footnote{\url{https://www.tcmtest.org.cn/}.}. This exam is an important test that evaluates candidates' theoretical knowledge, comprehension, and comprehensive application abilities in TCM. The exam content is mainly divided into the following 12 topics covering Basic TCM knowledge, classical literature, clinical TCM, basic Western medicine comprehensive, and basics in medical humanities:
\begin{itemize}

\item TCM basic theory, which covers basic concepts such as Yin-Yang, Five Elements, Zang-Fu theory, Qi, Blood, and Body Fluids. 

\item TCM diagnostics, including the four diagnostic methods: observation, listening and smelling, inquiry, and palpation, as well as the fundamental theories of differentiation and treatment.

\item Chinese Materia Medica, which studies the properties, channels, effects, and compatibility of Chinese medicinal herbs.

\item Formulae of TCM, which focuses on the composition, effects, indications, and applications of commonly used TCM formulas.

\item TCM classics, which cover content from classical TCM texts, including the "Huangdi Neijing" (Yellow Emperor's Inner Canon), "Shang Han Lun" (Treatise on Cold Damage), "Jingui Yaolue" (Essential Prescriptions of the Golden Cabinet), and Warm Diseases Theory, providing theoretical and clinical foundations for TCM practice.

\item TCM internal medicine, which studies the TCM diagnosis and treatment of internal diseases. 

\item TCM surgery, which studies the TCM diagnosis and treatment of surgical diseases.

\item TCM gynecology, which studies the TCM diagnosis and treatment of gynecological diseases.

\item TCM pediatrics, which studies the TCM diagnosis and treatment of pediatric diseases.

\item Acupuncture, which studies acupuncture treatment methods and their clinical applications.

\item Western medicine comprehensive, which assesses the basic knowledge of Western medicine, clinical professional knowledge, and infectious disease knowledge required for clinical practice, including basics of diagnostics, internal medicine (not tested for apprentice or specialty practitioners), and infectious diseases. 

\item Medical humanities, which assesses the legal regulations and ethical knowledge necessary for clinical practice, including Medical Ethics and Health Laws. 

\end{itemize}

We include multi-choice questions collected from publicly available TCM qualification examination question sets provided by \cite{yue2024tcmbench}, released under the CC BY 4.0 license, allowing for converting the original exam questions into a new format.

\textbf{Question Collection.} Multi-choice questions are gathered from the above sources. To ensure the diversity of topics and difficulty levels, we select 100 questions from the above 12 topics. The questions are then formatted into a standardized multi-choice format. Each question includes a stem (the main question), several distractors (incorrect options), and one correct answer. This formatting aligns with the structure commonly used in medical board exams. We then convert this structured format into a text sequence, allowing LLMs to read the questions.

\begin{figure*}
\centering
\includegraphics[width=0.84\textwidth]{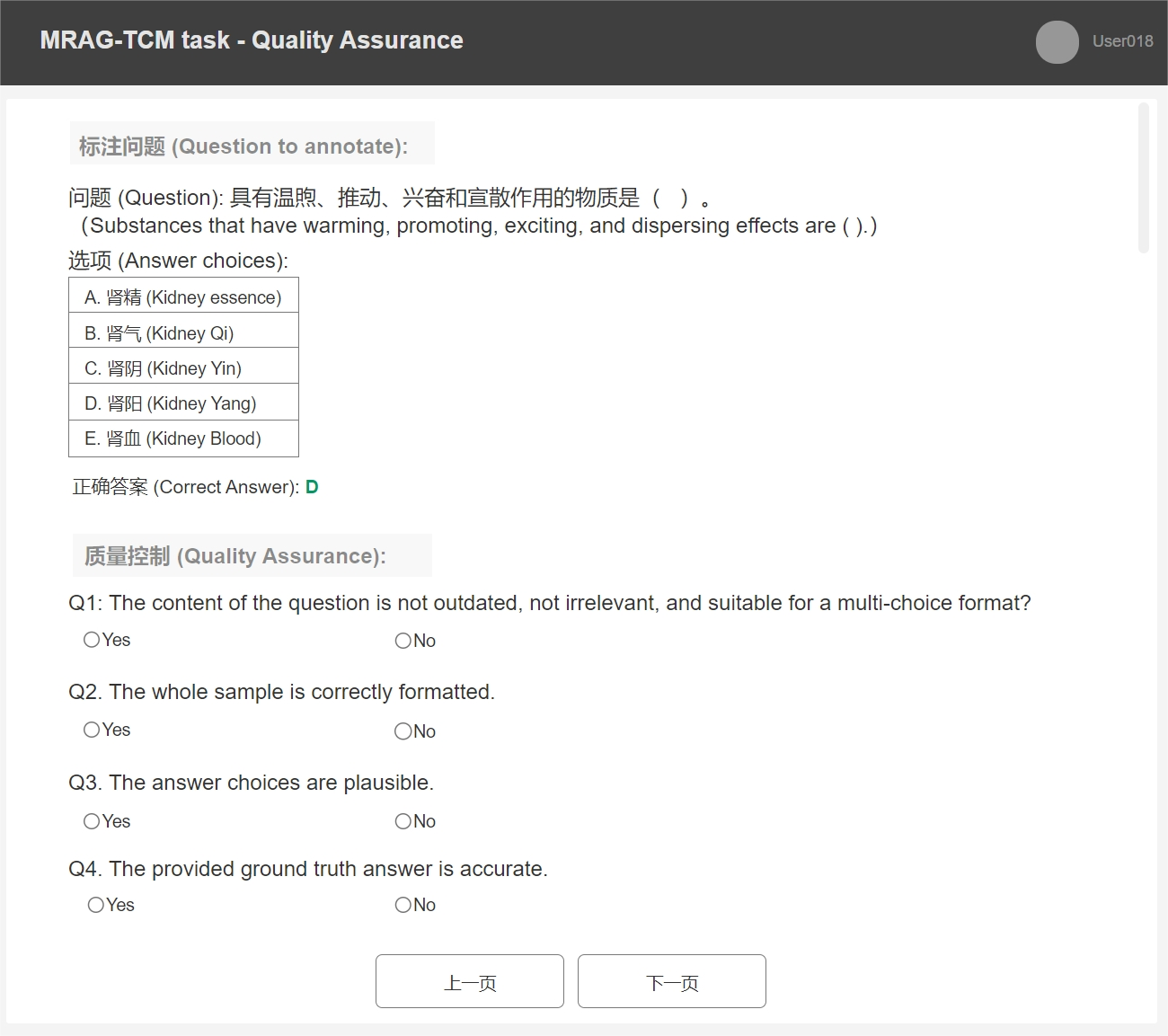} 
\caption{The screenshot of the annotation webpage for quality assurance of the MRAG-TCM tasks. }
\label{fig:screen_shot_tcm_annotation}
\end{figure*}

\textbf{Quality Assurance.} The dataset undergoes a review process for quality assurance. A pool of 15 TCM medical experts from China is divided into five groups, each containing three experts. These annotators participate in this project as volunteers and are paid 60 RMB per hour. A group will be given a TCM exam question in the multi-choice format.  They will check whether (a) the content of the question is not outdated, irrelevant, or not suitable for a multi-choice format, (b) the whole sample is correctly formatted, and (c) the answers are accurate. The screenshot of the annotation webpage is presented in Figure \ref{fig:screen_shot_tcm_annotation}. If not all of the experts in the group agree that the question is valid in all the above three aspects, this question will be filtered out. If a question is filtered out, another question with the same TCM topic is sampled from the sources and will go through the same review process. 

\textbf{Dataset Compilation.} After the quality assurance process, the questions are compiled into a structured dataset. Each entry in the dataset includes the question text, the possible answer choices, and the correct answer label.

\subsection{Curation of MRAG-CLFQA}

For the Chinese long-form question-answering task, we collected questions suitable for the retrieval augmented generation (RAG) system.

\textbf{Source.} For the Chinese long-form question-answering task, we collect 1,943 user queries from an online medical consultation platform\footnote{Due to the company policy, the name of the online medical consultation platform will be revealed upon acceptance.}. Each user is prompted to consent to data collection when collecting the queries. Furthermore, we ensure that no personal information is included in the dataset reviewing step. 

\begin{figure*}
\centering
\includegraphics[width=0.84\textwidth]{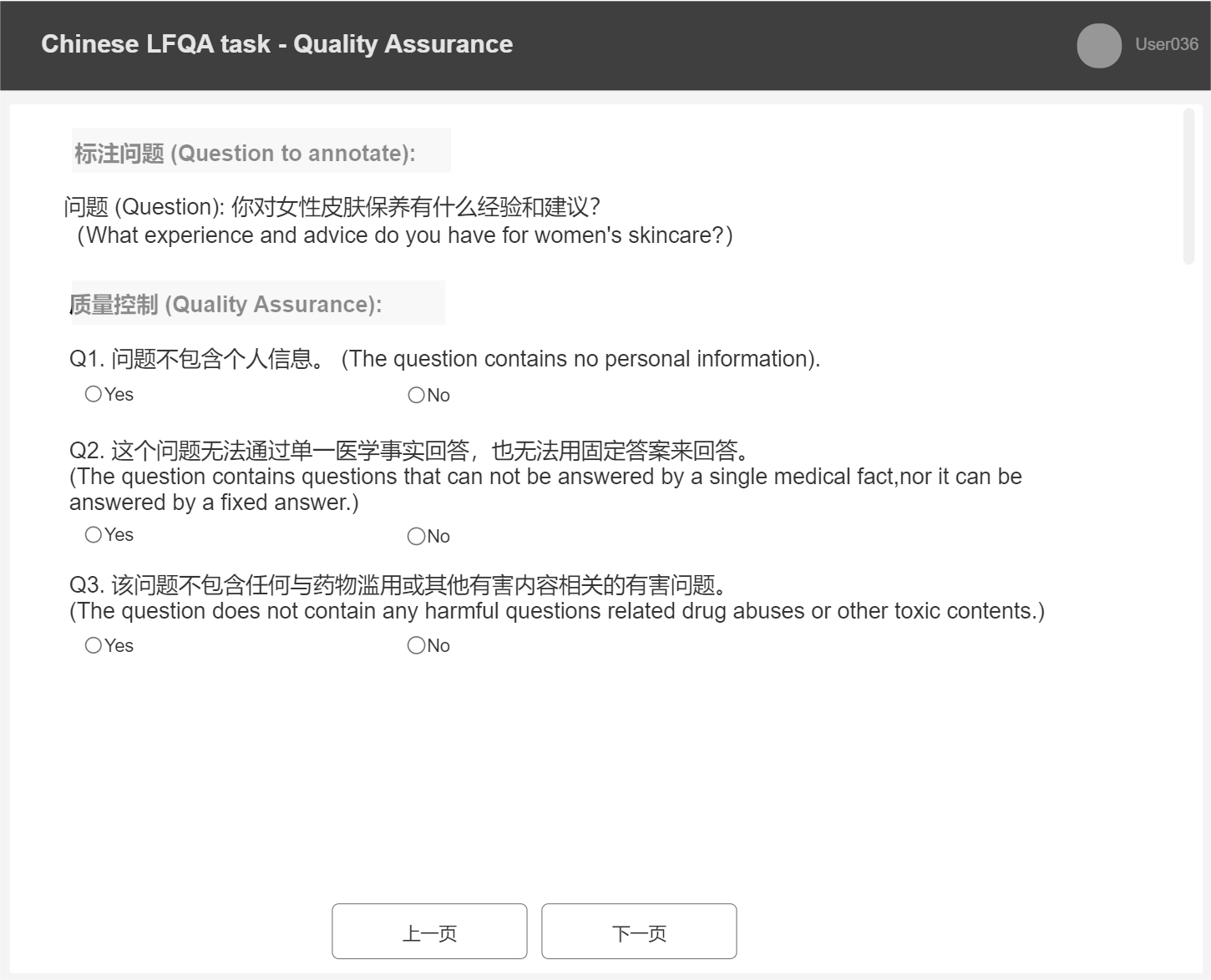} 
\caption{The screenshot of the annotation webpage for quality assurance of the MRAG-CLFQA tasks. }
\label{fig:screen_shot_clfqa_annotation}
\end{figure*}

\textbf{Dataset collection and filtering.} The collected dataset covers a wide range of topics, including (a) Symptoms, diagnosis, and treatment of illnesses. (b) prevention, vaccinations, or quarantine for infectious diseases. (c) lifestyle and wellness advice. A pool of 15 medical experts from China with medical doctoral degrees is divided into five groups, each containing three experts. These experts participate in this project as volunteers and are paid 10 US dollars per hour. Each group is randomly assigned a question, and the experts will check whether (a) the question contains no personal information. (b) a single medical fact can not answer the question. Moreover, one should refer to multiple documents to organize a proper response to the question. (c) The question does not contain any harmful questions related to drug abuse or other toxic content. The screenshot of the annotation webpage is presented in Figure \ref{fig:screen_shot_clfqa_annotation}. The question will not be included if the three experts do not unanimously agree upon any of the above aspects. Moreover, the remaining dataset contains 1253 medical queries.

\textbf{Formatting.} The medical queries are organized as a list of samples containing the query ID and the query's text string.

\subsection{Dataset statistics}

To summarize the datasets of the MRAG-Bench, we present the statistics of the datasets in Table \ref{tab:data_stats}. To ensure the balance of the dataset composition, the sizes of the included datasets will not exceed 1,500 samples.

\begin{table*}[tb!]
\centering
\resizebox{0.8\textwidth}{!}{
\begin{tabular}{cccccc}
\hline
\textbf{Dataset}  &   \textbf{Size}  &     \textbf{\#Options}     &    \textbf{Avg. Length}   &    \textbf{Task type}    &  \textbf{Language}    \\
\hline

MMLU-Med    &   1,089      &    4      &    63.5     &     MCQA   &  English     \\
MedQA-US    &   1,273      &    4     &    177.3   &   MCQA    &  English       \\
MedMCQA     &   1,500      &    4     &    26.1   &   MCQA     &  English            \\
PubMedQA     &   500       &    3     &    24.5   &   MCQA     &  English              \\
BioASQ       &     618      &   2     &    17.7   &   MCQA     &  English              \\ 
MRAG-TCM     &    1,200    &    4     &    29.8   &   MCQA     &  Chinese              \\ 

\hdashline
ChemProt        &     800    &    -   &    384.2   &   IE      &  English   \\ 
DDI            &    1,017     &    -   &       299.7   &   IE     &  English      \\ 
CMeIE           &   1,500     &    -   &       293.7   &   IE      &  Chinese     \\ 

\hdashline

ADInt            &    1,500   &    -   &      22.9   &   LP    &  English     \\  
DRKG            &     1,500   &     -   &      28.2   &   LP   &  English      \\  

\hdashline

MultiMedQA    &  1,066   &    -   &        10.2   &   LFQA  &  English   \\  
MRAG-CLFQA       &  1,253   &    -   &       70.9   &   LFQA  &  Chinese   \\  

\hline
\end{tabular}}
\caption{Statistics of MRAG-Bench tasks. \#Options: numbers of options; Avg. Length: average token counts in each question.  }
\label{tab:data_stats}
\end{table*}

\subsection{Evaluation metrics}
\label{subsec:app_evaluation_metrics}

We use objective evaluation metrics for the MCQA, IE, and LP task cohorts. We use post-processing scripts to transform the LLM's responses to structured data formats and use the following metrics: (a) we calculate the accuracy scores for the multi-choice questions. (b) The LP tasks use the same metric as MCQA. (c) For the IE tasks, we adopt the instance-level strict micro-F1 \cite{PromptCBLUE}; that is, the model predicts a triplet correctly if and only if it correctly predicts all its components.

For the LFQA tasks, we conducted a series of model- and human-based evaluations to assess the performance of LLM-based RAG systems. Since we are focusing on consumer health-related questions in which the audience is usually a layperson of average reading comprehension and no specific clinical context. Thus, we evaluate the LLM responses on the following aspects:
\begin{itemize}

\item \textbf{Usefulness}: The response should be helpful, safe, and informative and answer the query. 

\item \textbf{Readability}: The response should be well-organized and easy to follow for laypersons. 

\item \textbf{Knowledge}: The response should reflect the current consensus well and mention relevant and correct medical facts for answering the query. Moreover, no irrelevant information is discussed.

\item \textbf{Reasoning}: The response presents clear, correct reasoning steps.

\end{itemize}

In this work, LLMs (w/o. or w. RAG) are evaluated and ranked via pairwise matches over the above four axes. Our evaluation protocol is similar to Chatbot Arena \cite{chiang2024chatbot,zheng2024judging}.

\subsection{More information}

\textbf{Hosting.} \quad The MRAG-Bench's datasets are hosted in \url{https://huggingface.co/datasets/michaelwzhu/MRAG}. This URL is a permanent link and can be accessed by the community. 

\textbf{license.} The MRAG-Bench is released under the Creative Commons Attribution (CC BY 4.0) license.

\textbf{Author Responsibility Statement.} The authors of this dataset bear all responsibility for its content. The dataset has been created and shared to provide accurate and valuable data for research purposes. However, the authors do not assume liability for any errors, omissions, or inaccuracies in the dataset.

\textbf{User Responsibility.} By using this dataset, users agree to:
\begin{itemize}
\item Properly attribute the authors in any derivative works or publications that utilize the dataset.
\item Comply with all applicable laws and regulations, including data privacy and intellectual property rights.
\item Do Not use the dataset for any unlawful or unethical purposes.

\end{itemize}

The authors reserve the right to update or modify the dataset and its terms of use at any time. Users are encouraged to review the dataset and license periodically to ensure compliance with the current terms.

If anyone has any questions or requires further clarification regarding the use of this dataset, please contact \textit{wzhu91@connect.hku.hk}.

\section{Detailed Descriptions of MRAG-Toolkit}
\label{sec:app_mrag_toolkit}

In the main contents, we introduce the MRAG Toolkit to comprehensively evaluate how different LLM-based RAG systems perform on our MRAG Bench. As shown in Figure 2, the MRAG Toolkit consists of three major components: corpora, retrievers, and response generators. We now introduce these components in detail.

\subsection{Corpora} 
\label{subsec:app_mrag_toolkit_corpora}

In this work, we utilize four different corpora for the English tasks: the medical corpus, the open-domain corpus, the combined corpus, and the World Wide Web. The combined corpus is the combination of the first two. The medical corpus is the combination of the following resources for medical literature or textbooks:
\begin{itemize}
\item PubMed\footnote{\url{https://pubmed.ncbi.nlm.nih.gov/}} is the most widely used literature resource, containing millions of biomedical articles. Many relevant studies use PubMed as the retrieval corpus \cite{frisoni2022bioreader,naik2021literature}. We use a PubMed subset of 23.9 million articles with valid titles and abstracts for this work. The processed dataset is available at \url{https://huggingface.co/datasets/michaelwzhu/MRAG/tree/main}. 

\item StatPearls\footnote{\url{https://www.statpearls.com/}} is a point-of-the-care clinical decision support tool similar to UpToDate\footnote{\url{https://www.wolterskluwer.com/en/solutions/uptodate}}. We use the 9,330 publicly available StatPearl articles through NCBI Bookshelf14 to construct the StatPearls corpus. We chunked StatPearls according to the hierarchical structure, treating each paragraph in an article is a snippet, and all the relevant hierarchical headings are spliced as the corresponding title. 

\item Textbooks\footnote{\url{https://github.com/jind11/MedQA}} is a collection of 18 widely used medical textbooks, which are important references for medical students taking the United States Medical Licensing Examination (USLME). In MRAG, the textbooks are processed as chunks with
no more than 1000 characters. We used the \textbf{RecursiveCharacterTextSplitter} from
LangChain\footnote{\url{https://github.com/langchain-ai/langchain}} to perform the chunking. 

\end{itemize}

For constructing an open-domain corpus, we utilize the Wikipedia (English) corpus\footnote{\url{https://en.wikipedia.org/wiki/Wikipedia:Database_download}}. As a large-scale open-source encyclopedia, Wikipedia is frequently used as a corpus
in information retrieval tasks \cite{thakur2021beir} and open-domain question-answering tasks \cite{chen2017reading}. We select Wikipedia as one of the corpora to see if the general domain database can be used to improve the ability of medical QA. We also chunked Wikipedia's documents with LangChain.

The World Wide Web can also serve as an extensive and dynamic retrieval corpus for Retrieval-Augmented Generation (RAG), offering a vast and diverse repository of information across virtually all knowledge domains. Leveraging the web as a retrieval corpus enables RAG systems to access up-to-date content, providing rich context and comprehensive data sources for generating accurate and relevant responses. This expansive corpus includes various formats, from scholarly articles and news reports to blogs, forums, and multimedia content, ensuring a breadth of perspectives and insights. The web's continuously evolving nature could enhance the RAG system's ability to produce informed and current outputs. It is an invaluable resource for applications requiring real-time information retrieval and generation. However, the web page contents could also introduce noise or false information to the RAG system. In this work, we utilize the Bing Search API\footnote{\url{https://www.microsoft.com/en-us/bing/apis/bing-web-search-api}} to access and retrieve relevant documents from the web.

For the Chinese tasks, we utilize (a) the World Wide Web and (b) a proprietary medical corpus and open-domain corpus owned by a company. The company’s name and detailed information on the corpus will be revealed upon acceptance.

To summarize the retrieval corpora, we present their statistics in Table \ref{tab:corpus_stats}.

\begin{table*}[tb!]
\centering
\resizebox{0.96\textwidth}{!}{
\renewcommand\arraystretch{1.3}
\begin{tabular}{ccccccc}
\hline
\textbf{Corpus}  &  \textbf{Source}    &    \textbf{\#Docs}  &     \textbf{\#Snippets}     &    \textbf{Avg. Length}   &    \textbf{Domain}    &  \textbf{Language}    \\
\hline

\multirow{3}*{Medical corpus}    &   PubMed     &   23.9M   &     23.9M   &     296   &   Biomedicine     &  English    \\
&    StatPearls  &  9.3k  &    301.2k  &  119   &   Clinics   &  English    \\

&   Textbooks   &    18  &    125.8k   &     182  &   
  Medicine &  English  \\

\hdashline

Open-domain corpus    &    WikiPedia    &   6.5M 
   &     29.9M     &   162    &    General     &   English    \\

\hdashline

World wide web   &    Internet    &   - 
   &     -    &   -   &    General     &    English \& Chinese    \\

\hdashline

Proprietary medical corpus   &    proprietary    &   3.6M 
   &      13.5M    &   348   &    Medical     &   Chinese    \\

\hline
\end{tabular}}
\caption{Statistics of the retrieval corpora. \#Docs: the number of documents contained in the corpus. \#Snippets: the number of document snippets contained in the corpus. Avg. Length: average token counts in each document snippets. }
\label{tab:corpus_stats}
\end{table*}

\subsection{Retrievers} 

In this work, we consider the following retrievers for the English MRAG-Bench tasks:
\begin{itemize}
\item Best Matching 25 (BM25) \cite{robertson2009probabilistic}. BM25 is a highly effective lexicon-based sparse retrieval algorithm commonly utilized for information retrieval tasks, such as in Retrieval-Augmented Generation (RAG) for large language models. BM25 scores the relevance of documents by considering the frequency and distribution of query terms within those documents. Specifically, it enhances traditional term frequency-inverse document frequency (TF-IDF) methods by incorporating term saturation and document length normalization. BM25 ensures that the relevance score increases logarithmically with term frequency, avoiding excessive influence from overly common terms, and adjusts for document length to prevent bias toward longer documents. By weighing query terms according to their inverse document frequency and accounting for term saturation, BM25 provides a robust and scalable approach for retrieving pertinent documents in RAG, enhancing the contextual accuracy and informativeness of the generated responses.

\item MedCPT \cite{jin2023medcpt}. MedCPT is a biomedical embedding model pre-trained with contrastive loss on 255 million user clicks from PubMed search logs. It achieved state-of-the-art performance on several biomedical IR tasks. For our experiments, we use the embedding model to transform the document snippets to vectors and build a vector index with the help of Faiss\footnote{\url{https://github.com/facebookresearch/faiss}}. Upon receiving a user query, the embedding model embeds the query to a vector and leverages the efficient nearest neighbor search techniques (also implemented in Faiss) on vectors. Vector-based search is highly efficient since a search can be done in 3 ms with a vector index of sizes in billions.

\item BGE-base \cite{xiao2023c}. The BGE-base model is a sophisticated sentence embedding model designed to transform sentences into high-dimensional vector representations, enabling efficient and meaningful comparison of textual data. This model leverages pre-training on large-scale corpora to deeply understand language and provide high-quality semantic representations for input documents. 

\item E5-Mistral-7B \cite{wang2023improving}, a LLM based retriever. This model uses the Mixtral-7B as the document encoder and is further pre-trained on a large-scale synthetic dataset via the contrastive learning loss function.

\item RRF \cite{cormack2009reciprocal} proposed to merge results from different retrievers with Reciprocal Rank Fusion (RRF), which effectively fuses the information from different sources by selecting shared predictions. In this work, we utilize this approach to combine results from BGE-base and MedCPT. 

\end{itemize}

For the Chinese tasks, we utilize the BGE-base Chinese model\footnote{\url{https://huggingface.co/BAAI/bge-base-zh}} as the retriever, if the local corpus is used.

We summarize the basic information of the retrievers in Table \ref{tab:retriever_stats}.

\begin{table*}[tb!]

\centering
\resizebox{0.7\textwidth}{!}{
\renewcommand\arraystretch{1.3}
\begin{tabular}{cccccc}
\hline
\textbf{Retriever}  &  \textbf{Type}    &    \textbf{Size}  &     \textbf{Metric}     &    \textbf{Domain}   &   \textbf{Language}    \\
\hline

BM25    &   lexical     &   -   &   BM25  &      General 
  &    -   \\

MedCPT    &   Semantic     &    109M   &      cosine similarity     &    Biomedicine   &  English   \\

BGE-base    &   Semantic     &    110M   &   cosine similarity     &    General   &  English   \\

E5-Mistral-7B  &   Semantic     &    7B   &   cosine similarity     &    General   &  English   \\

BGE-base Chinese  &   Semantic     &    110M   &   cosine similarity     &    General   &  Chinese   \\

\hline
\end{tabular}}

\caption{Statistics of the retrievers.  }
\label{tab:retriever_stats}

\end{table*}

\subsection{LLMs as response generator} 

In this work, we select the most frequently used or recently released LLMs with excellent performance in the open-domain evaluation benchmarks to evaluate RAG systems. 
\begin{itemize}
\item Commercial LLMs developed by the OpenAI, GPT-3.5 (gpt-3.5-turbo), and GPT-4 (gpt-4o). These two models are popular commercial LLMs, which have already shown great capabilities in directly answering medical questions \cite{singhal2023large,nori2023capabilities}. We access these two models via the APIs provided by OpenAI\footnote{\url{https://platform.openai.com/docs/models}}.

\item The Tongyi Qwen (qwen\_max) model\footnote{\url{https://tongyi.aliyun.com/qianwen/}} is an advanced language model developed by Alibaba Cloud, designed to push the boundaries of natural language processing and generation capabilities. This model, built upon extensive datasets and cutting-edge deep learning algorithms, aims to excel in a wide range of tasks in both Chinese and English, including text generation, conversation, summarization, translation, and more.  

\item The Mixtral-8x22B (-Instruct-v0.1) model\footnote{\url{https://mistral.ai/news/mixtral-8x22b/}} is one of the latest open-sourced LLM. It sets a new standard for performance and efficiency within the AI community. It is a sparse Mixture-of-Experts (SMoE) model that uses only 39B active parameters out of 141B, offering unparalleled cost efficiency for its size. Mixtral-8x22B has the following strengths: (a) It is fluent in English, French, Italian, German, and Spanish. (b) It has strong mathematics and coding capabilities. (c) It is natively capable of function calling. (d) Its 64K tokens context window allows precise information to be recalled from large documents. This model is released under Apache 2.0, the most permissive open-source license, allowing anyone to use the model anywhere without restrictions.

\item Qwen2.5 \cite{bai2023qwen} is a language model series including decoder language models of different sizes. It is based on the Transformer architecture with SwiGLU activation, attention QKV bias, group query attention, a mixture of sliding window attention and full attention. Additionally, it has an improved tokenizer that is adaptive to multiple natural languages and codes. In this work, unless otherwise specified, we use the Qwen-1.5-72B (-chat) \footnote{\url{https://huggingface.co/Qwen/Qwen2.5-72B}} model. 

\item Meta developed and released the Meta Llama 3 family\footnote{https://llama.meta.com/llama3/} of large language models (LLMs), a collection of pretrained and instruction-tuned generative text models in 8 and 70B sizes. The Llama 3 instruction-tuned models are optimized for dialogue use cases and outperform many of the available open-source chat models on standard industry benchmarks. Further, in developing these models, Meta significantly optimized helpfulness and safety. Unless otherwise specified, we use the Llama-3-70B (-Instruct) model in this work.

\item MEDITRON \cite{chen2023meditron} is a series of biomedical LLMs built based on Llama2 \cite{Touvron2023Llama2O} and fine-tuned on open-source biomedical literature. In this work, we use its 70B version model. 

\item PMC-LlaMA (13B) \cite{wu2023pmc} is fine-tuned based on LLaMA \cite{Touvron2023Llama2O}, using the medical literature from PubMed.

\end{itemize}

For the Chinese tasks, the following LLMs will be evaluated: 
\begin{itemize}
\item GPT-3.5 (gpt-3.5-turbo).

\item GPT-4 (gpt-4o). 

\item Tongyi Qwen (qwen\_max).

\item Qwen-1.5 72B.

\item DISC-MedLLM \cite{bao2023disc} leverages Large Language Models (LLMs) to provide accurate and truthful medical responses in end-to-end conversational healthcare services. It constructs high-quality Supervised Fine-Tuning (SFT) datasets by utilizing medical knowledge graphs, reconstructing real-world dialogues, and incorporating human-guided preference rephrasing. With the constructed high-quality dataset, DISC-MedLLM is fine-tuned from Baichuan-13B-Base\footnote{\url{https://huggingface.co/baichuan-inc/Baichuan-13B-Base}} model and surpasses many Chinese medical LLMs in both single-turn and multi-turn consultation scenarios. 

\end{itemize}

Unless otherwise specified, all the LLMs utilize the nucleus sampling strategy \cite{holtzman2019curious} for decoding. The temperature parameter is set to 0.7, and the top\_p parameter is set to 0.8.

We summarize the basic information of the LLM response generators in Table \ref{tab:llm_stats}.

\begin{table*}[tb!]

\centering
\resizebox{0.9\textwidth}{!}{
\renewcommand\arraystretch{1.3}
\begin{tabular}{cccccc}
\hline
\textbf{LLM}  &  \textbf{Size}    &    \textbf{Context size}  &     \textbf{Open-source}     &    \textbf{Domain}   &   \textbf{Language}    \\
\hline

GPT-3.5   &      -   &    16,385   &    False   & General     &      English \& Chinese       \\
GPT-4   &      -   &  128,000   &    False        & General     &       English \& Chinese         \\
Tongyi Qwen   &       -   &     8,000   &     False      & General     &        English \& Chinese           \\
Mixtral-8x22B    &   141B (39B activated)  &    65,536   
 &     True      &    General     &    English       \\ 
Qwen2.5-72B   &  72B    &     32,768    &     True      &    General     &     English \& Chinese          \\

LlaMA-3-70B   &    70B   &   8,192    &    True      & General     &      English       \\
MEDITRON    &   70B    &     4,096     &     True      &    Biomedicine     &    English      \\ 
PMC-LlaMA    &  13B     &   4,096     &    True    &    Biomedicine     &     English   \\

DISC-MedLLM    &  13B     &   4,096    &    True    &   Medicine     &    Chinese    \\ 

\hline
\end{tabular}}
\caption{Statistics of the LLM response generators. }
\label{tab:llm_stats}

\end{table*}

\subsection{Prompting strategies}
\label{subsec:prompting_strategies}

We now describe the prompting strategies used when evaluating the LLMs on the MRAG bench. Following \cite{xiong2024benchmarking}, all the RAG systems should be evaluated in a zero-shot setting where in-context few-shot learning is not permitted.

The prompting strategy can be classified as either (a) w/o. RAG or (b) w. RAG, based on whether an LLM retrieves referential documents and concatenates them to the prompt. In this work, we only consider the framework where the retrieved knowledge/contents are concatenated in the prompt, and no other approaches, like memory augmentation, are applied to insert external information into the LLMs. 

Based on how the response is elicited, the prompting strategy can be classified as: (a) Direct answer (DA): given the question, the prompt asks the LLM to output the answer directly. (b) Chain-of-thought (COT) \cite{Wei2022ChainOT} explicitly asks the LLM to think step by step and demonstrate the intermediate outputs. (c) COT-Refine. Building on COT and Self-Refine\cite{madaan2024self}, we developed a simple prompting strategy called COT-refine. This strategy involves a two-stage process: first, given a COT prompt and a question, the model produces a response (R0). Then, in the second stage, the model is conditioned on the original prompt, question, and R0 and is prompted to produce a refined answer with detailed explanations. This strategy allows the LLM to reflect on the previous answer and make necessary corrections. Regarding the prompt strategy for eliciting responses, COT-Refine is used by default.

\end{CJK*}

\end{document}